\documentclass[a4]{article}
\usepackage[affil-it]{authblk}

\usepackage{amsfonts, amstext, amssymb}
\usepackage{amsmath, amsthm}
\usepackage{xspace}

\usepackage{siunitx}
\usepackage{subcaption}
\usepackage{algorithmicx}
\usepackage{algpseudocode}
\usepackage{algorithm}
\usepackage{hyperref}
\usepackage{bookmark}

\usepackage{orcidlink}

\usepackage{pgfplots}
\usepgfplotslibrary{fillbetween}

\usepackage{tikz,ifthen}
\usepackage{tikz-3dplot}

\pgfplotsset{compat=1.18}

\newtheorem{theorem}{Theorem}
\newtheorem{lemma}[theorem]{Lemma}
\newtheorem{corollary}[theorem]{Corollary}

\newtheorem{example}{Example}

\def\IR{\textit{IR}\xspace}
\def\IA{\textit{IA}\xspace}
\def\defeq{\stackrel{\text{\tiny{}def}}{=}}

\newcommand{\Mdist}[2]{\left\|{#1}-{#2}\right\|}

\newcommand{\interior}[1]{%
  ({\kern0pt#1})^{\mathrm{o}}%
}

\newcommand{\sumM}[1]{%
\sum{#1}}

\begin{document}

\title{Significativity Indices for Agreement Values}

\author{Alberto Casagrande}
\affil{Dept.~of Mathematics, Computer Science, and Physics\\ University of Udine}
\author{Francesco Fabris}
\affil{Dept~of Mathematics, Computer Science, and Geosciences\\ University of Trieste}
\author{Rossano Girometti}
\affil{Dept.~of Medicine\\ University of Udine}
\author{Roberto Pagliarini}
\affil{Dept.~of Mathematics, Computer Science, and Physics\\ University of Udine}

\date{July 3, 2025}

\maketitle

\textit{This is a draft version of the article published on ``\emph{Statistics and Computing}''.
The final version is available at \href{https://doi.org/10.1007/s11222-025-10728-1}{https://doi.org/10.1007/s11222-025-10728-1}.}

\begin{abstract}
Agreement measures, such as Cohen's kappa or intraclass correlation, gauge 
the matching between two or more classifiers.  They are used in a wide range of contexts from medicine, where they evaluate the effectiveness of medical treatments and clinical trials, to artificial intelligence, where they can quantify the approximation due to the reduction of a classifier.
The consistency of different classifiers to a golden standard can be compared simply by using the order induced by their agreement measure with respect to the golden standard itself. Nevertheless, labelling an approach as good or bad exclusively by using the value of an agreement measure requires a scale or a significativity index. Some quality scales have been proposed in the literature for Cohen's kappa, but they are mainly na\"ive, and their boundaries are arbitrary.
This work proposes a general approach to evaluate the significativity of any agreement value between two classifiers and introduces two significativity indices: one dealing with finite data sets, the other one handling classification probability distributions. Moreover, this manuscript addresses the computational challenges of evaluating such indices and proposes some efficient algorithms for their evaluation.
\end{abstract}

\section{Introduction}

Classifiers are processes that label entries in a data set. They may be fully automated, such as algorithms (e.g., see~\cite{726791,7780459}), or require human activities, such as in clinical evaluations (e.g., see~\cite{PIRADS2,birads2013}). The ideal (or perfect) classifier is the one that correctly labels the entries according to the \emph{golden standard}: a labelling that represents the highest and unquestionable knowledge about the domain. In machine learning, the golden standard corresponds to training and test data set labelling. In the clinical context, it is the definition of the investigated disease or condition which may correspond to the result of a clinical exam; for example, hypoglycemia is diagnosed when the result of the fasting blood glucose test is below \SI{70}{\milli\gram\per\deci\liter}~\cite{Elamin2024}. 

When resources are limited, the ideal classifier may be non-feasible, and alternatives must be considered. For instance, PET-CT (\emph{Positron Emission Tomography-Computed Tomography}) is widely used to detect, stage, and monitor various types of cancer (e.g., see~\cite{MA2025110852}), but it is not suggested in the screening of low-risk subjects due to its costs and drawbacks. Analogously, quantised neural networks are valid alternatives to their plain counterparts to gain efficiency or even low-resource hardware evaluability at the price of some accuracy~\cite{hinton2015distillingknowledgeneuralnetwork}.

Agreement measures have been introduced since the XX century to measure differences among classifiers.
Some of them gauge the agreement between pairs of classifiers, such as log odds ratio~\cite{dda9a3b2-7e66-38cd-8ce8-7d36cd60b878}, McNemar’s test~\cite{McNemar:1947}, Cohen’s kappa~\cite{Cohen:1960} and its multiclass generalization~\cite{Cohen:1968}, intraclass correlation~\cite{Shrout:1979}, and information agreement ($\IA$)~\cite{Casagrande:2020a, Casagrande:2020b, Casagrande:2022, Casagrande:2023}. Others deal with sets of classifiers such as Fleiss's kappa~\cite{Fleiss71}, the adjusted rand index~\cite{hubert1985comparing}, the consensus clustering~\cite{monti2003consensus}, and Krippendorff's alpha~\cite{krippendorff2004reliability}. Most of the agreement measures report the agreement values in the real interval $[-1,1]$ where $-1$ means total disagreement, while $1$ is associated with complete agreement. Information agreement instead rates the agreement in the interval $[0,1]$ because this measure quantifies the information exchanged by two classifiers during the classification process~\cite{Casagrande:2022}.

Agreement measures are usually used to order the performances of different classifiers with respect to the perfect one. If the agreement of a classifier $C_1$ with the ideal one is higher than that of the classifier $C_2$, then $C_1$ is preferable to $C_2$ in terms of performance. However, the meaningfulness of the agreement values by themselves, i.e., the values reported by the agreement measures, is not easily interpretable, and their significativity may be obscure. How much do two classifiers with Cohen's kappa $0.7$ agree? Is $0.7$ a significant value? 
In order to address such questions, Landies and Koch proposed a linear scale to interpret the strength of the agreement based on Cohen's $\kappa$: $[0, 0.2)$ (``none
to slight''), $[0.2, 0.4)$ (``fair''), $[0.4, 0.6)$ (``moderate''),
$[0.6, 0.8)$ (``substantial'') and $[0.8, 1.0)$ (``perfect or almost perfect agreement'')~\cite{Landies1977}. 
%However, this interpretation allows for very little agreement among raters to be described as ``substantial''.
This scale considers the $0.61$ agreement to be ``substantial''. A different scale suggests that values greater than $0.75$ represent excellent agreement beyond chance, values in the interval $[0.40 , 0.75]$ correspond to fair to good agreement,  and values below $0.40$ are poor agreement~\cite{fleiss2013statistical}.
Therefore, these scales are either missing or, in the best case, totally arbitrary.

This work deals with agreements among pairs of classifiers. 
In such cases, the joint behaviour of two classifiers can be summarised by 
a \emph{confusion matrix} or a \emph{probability matrix}. The cell $(i,j)$
in the former kind of matrices stores the number of entries in the data set that are labelled 
as belonging to the $i$-th class by the first classifier and to the $j$-th class by
the second classifier. Instead, the cell $(i,j)$
in a probability matrix contains the joint probability for an entry to be labelled as
belonging to the $i$-th class and, at the same time, to the $j$-class by the first and the 
second classifiers, respectively.
Thus, any agreement measure is a function from the set of confusion or probability matrices to the set of agreement values.

In this context, we introduce two significativity indices for agreement values: 
one on the confusion matrices and the other one on the probability matrices. 
Both of them define the significativity of
an agreement value obtained over a data set as
the probability to randomly select a matrix built
over the same data set with a lower agreement value.
%This induces a parametric scale on the considered agreement measure and relates any value to its probability.
%For example, if $75\%$ of the classifier pairs have a Cohen's $\kappa$ of lower than $0.281567$, the significativity of $0.281567$ with respect to Cohen's $\kappa$ will be $0.75$.
%and placed at one fourth of the $\kappa$ significativity scale.
The more likely it is to choose a matrix with a lower agreement value, the higher the
significativity of the agreement.
From another point of view, the more difficult it is to find two classifiers whose agreement is 
at least equivalent to the investigated value, the higher its significativity.

The proposed indices are parametric in both the investigated agreement measure and 
the number of classes. The index on the confusion matrices also depends on the size of the original data set.
This approach is analogous to the classical statistical coefficient $p$-value~\cite{fisher1925statistical}, which measures the probability for the null hypothesis to comply with the data. As long as the $p$-value is under a selected threshold -- usually $0.05$ -- the null hypothesis is discarded. Our method does not set any threshold, leaving the task of identifying one to the user. Instead, it evaluates the probability for a random matrix to have an agreement value lower than the considered one to measure its significativity. 

The proposed indices do not gauge the meaningfulness of the original data set, but exclusively deal with the agreement values. The same agreement value can be obtained from two data sets with substantially different cardinalities, and the agreement value of a confusion matrix may have a high significativity even though the data set used to build it consists of a few entries. Thus, we can deduce that agreement and data set significativities are not directly related.

Our method has three main advantages. % over those available in the literature. 
First, it associates any agreement measure, even those whose meaning is obfuscated, 
with a clear and consolidated index: the probability of decreasing the agreement value by chance. 
Second, the induced scale is not arbitrary, as it reports the probabilities of the agreement values, and it deals with objective quantities. Finally, this method may help compare agreement measures, providing a familiar and unifying approach.

%The resulting scales are not equivalent because they are parametric on the original agreement measures, which establish the quality order among the matrices. 

This work is organised as follows: Section~\ref{sec:notions}
introduces the basic notions and notation.
Section~\ref{sec:significativity_confusion} defines the notion of 
$\sigma$-significativity over confusion matrices, where $\sigma$ is any agreement
measure between two classifiers, it studies the asymptotic time complexity 
of computing this value, and it proposes a time-feasible 
numeric estimator for it.
Section~\ref{sec:significativity_probability} introduces the $\sigma$-significativity over 
probability matrices and proposes an efficient algorithm to numerically
estimate it. The same section also 
proves that, under some reasonable conditions on the syntactic form 
of $\sigma$, the $\sigma$-significativity over confusion
matrices converges to the $\sigma$-significativity over probability matrices 
as the size of the confusion matrix data set tends to infinity. 
Finally, Section~\ref{sec:conclusions} summarises the achieved results, contains the 
concluding remarks, 
and suggests some possible future developments.

\section{Basic Notions and Notation}\label{sec:notions}

For any set $S$ and for any pair of positive natural values $n, m \in \mathbb{N}_{>0}$, the set of $n \times m$ matrices with elements in $S$ is denoted by $S^{n \times m}$. If $M$ is a matrix, then $M(i,j)$ is 
$M$'s element in the $i$-th row and the $j$-th column. We may write $\sumM{M}$ meaning the sum of the 
elements in $M$, i.e., when $M \in S^{n \times m}$, $\sumM{M}\defeq\sum_{i=1}^n\sum_{j=1}^m M(i,j)$.

If $f: A \rightarrow B$ and $C \subseteq A$, we may write $f(C)$ meaning the image set of $C$, i.e., $f(C)=\{f(c)\,|\, c \in C\}$.

For any set $S$, the \emph{indicator function of $S$}, %$\mathbf{1}_{\mathcal{S}_{\sigma}(c)}(x): \mathbb{R}^{n^2} \rightarrow \{0,1\}$, 
\begin{equation}
\mathbf{1}_{S}(x) \defeq 
\begin{cases}
0 & \text{if $x \not \in S$}\\
1 & \text{if $x \in S$},
\end{cases}
%\begin{cases}
%0 & \text{if $x \not \in \Delta^{(n^2-1)} \lor \sigma(\gamma(x)) < c$}\\
%1 & \text{if $x \in \Delta^{(n^2-1)} \land \sigma(\gamma(x)) < c$},
%\end{cases}
\end{equation}
maps the elements of $S$ to $1$ and all the other elements to $0$.

A \emph{classifier} is a process or a rater that partitions the investigated domain $\mathcal{D}$ into 
$n$ distinct classes. It may be implemented as a digital system (e.g., AI models), a medical exam (e.g., COVID-19 test), or 
a clinical evaluation (e.g., BI-RADS or PI-RADS).
Any classifier corresponds to a function $\chi: \mathcal{D} \rightarrow [1,n]$. When $\chi(d)=i$, we say that \emph{$d$ is in the class $i$ according to $\chi$}.

Two classifiers $\chi_1$ and $\chi_2$ can be compared by evaluating $m \in \mathbb{N}$ distinct elements in $\mathcal{D}$.
The result is the evaluations can be collected in a \emph{$n\times n$ confusion matrix} consisting of $m$ tests that is a 
$n \times n$ matrix $M_C \in \mathbb{N}^{n\times n}$ whose value in position $(i, j)$ reports the number of the $m$ elements which are in the 
classes $i$ and $j$ according to $\chi_1$ and $\chi_2$, respectively, i.e., $M_C(i,j)=|\{d \in \mathcal{D}\,|\, \chi_1(d)=i \land \chi_2(d)=j\}|$.
For any $n\times n$ confusion matrix, $M_C$, built from a data set of size $m$, the sum of its values is $m$, i.e., %$m=\sum_{i=1}^n\sum_{j=1}^n M_C(i,j)$.
$m=\sumM{M_C}$.
 
The set of all the $n\times n$ confusion matrices built over $m$ tests is denoted by $\mathcal{M}_{n, m}$, i.e., 
%\begin{equation}
%\mathcal{M}_{n, m} \defeq \left\{\begin{pmatrix}c_{1}&\ldots&c_{n}\\ \vdots& \ddots&\vdots\\c_{n(n-1)+1}&\ldots& c_{n^2}\end{pmatrix} \in \mathbb{N}^{n \times n}\, |\, \sum_{i=1}^{n^2} c_i = m \right\}.
%\end{equation}
%
\begin{equation}
\mathcal{M}_{n, m} \defeq \left\{M_C \in \mathbb{N}^{n \times n}\, \big|\, \sumM{M_C} = m \right\}.
\end{equation}

A \emph{$n\times n$ probability matrix} is a real-valued matrix representing the joint probability distribution of a pair of 
independent and discrete random variables ranging in the interval $[1,n]$. 
The value in position $(i,j)$ is the probability that the first random variable returns $i$ and, at the same time, the second returns $j$. 
Since every probability matrix $M_P$ is a discrete probability distribution, its values are non-negative and sum up to $1$, i.e., 
%$1=\sum_{i=1}^{n} \sum_{j=1}^{n} M_P(i,j)$ 
$\sumM{M_P}=1$
and  $M_P(i,j) \geq 0$ for all $i \in [1,n]$ and for all $j \in [1,n]$.

The set of all the $n\times n$ probability matrices, $\mathcal{P}_n$, is defined as follows 
%\begin{equation}
%\mathcal{P}_n \defeq \left\{\begin{pmatrix}p_{1}&\ldots&p_{n}\\ \vdots& \ddots&\vdots\\p_{n(n-1)+1}&\ldots& p_{n^2}\end{pmatrix} \in \mathbb{R}^{n \times n}\, |\, \sum_{i=1}^{n^2} p_i = 1 \land \bigwedge_{i = 1}^{n^2} p_i \geq 0 \right\}.
%\end{equation}
\begin{equation}
\mathcal{P}_n \defeq \left\{M_P \in \mathbb{R}_{\geq 0}^{n \times n}\, \bigg|\, \sumM{M_P} = 1 \right\}.
\end{equation}

Every confusion matrix $M_C \in \mathcal{M}_{n,m}$ induces a probability matrix $M_P \in \mathcal{P}_n$ by means of the function $T_P$ defined as follows
%\begin{equation}\label{def:T_P}
%T_P\left(\begin{pmatrix}c_{1}&\ldots&c_{n}\\ \vdots& \ddots&\vdots\\c_{n(n-1)+1}&\ldots& c_{n^2}\end{pmatrix}\right)
%\defeq \frac{1}{\sum_{i=1}^{n^2}c_{i}} \begin{pmatrix}c_{1}&\ldots&c_{n}\\ \vdots& \ddots&\vdots\\c_{n(n-1)+1}&\ldots& c_{n^2}\end{pmatrix}.
%\end{equation}
\begin{equation}\label{def:T_P}
T_P\left(M_C\right) \defeq \frac{1}{\sumM{M_C}}M_C.
\end{equation}

A \emph{agreement measure} is a function $\sigma: \mathcal{P}_n \cup \bigcup_{m \in \mathbb{N}} \mathcal{M}_{n, m} \rightarrow I_{\sigma}$, where $I_{\sigma}$ is an interval over $\mathbb{R}$,
meant to 
measure the agreement between two classifiers or the effectiveness of a classifier with respect to a golden standard 
using either a confusion or a probability matrix. Cohen's $\kappa$~\cite{Cohen:1960}, Scott's $\pi$~\cite{10.1086/266577}, Yule's $Y$~\cite{Yule12}, 
Fleiss's $\kappa$~\cite{Fleiss71}, and %$\IR$~\cite{Girometti:2015}, and
$\IA$~\cite{Casagrande:2020a,Casagrande:2023,Casagrande:2024aa} are agreement measures.   

Later on, we will assume agreement-ordered agreement measures, meaning that greater agreement between the classifiers will correspond to 
greater agreement values. This condition can be easily relaxed when the lower values correspond to large agreement, for instance by considering the opposite of the investigated agreement measure. 
Nevertheless, this assumption simplifies the presentation of the formal aspects and the notation.

\subsection{O-Minimal Theories and Definability}~\label{sec:ominimal}

Section~\ref{sec:significativity_probability} relates some of the properties of the 
investigated agreement measures and the language used to define them.
Because of this, we need to introduce the notion of theory as a syntactic characterisation for
sets.

A \emph{theory} is a set of first-order formulas that describe a class of structures. Any theory
is defined by the variable domain, a set of constants, a set of functional symbols, and a 
set relational symbols. When the variables in the theory assume values in the set $\mathcal{Q}$, 
we say that the theory is over $\mathcal{Q}$.
The Presburger arithmetic theory, $(\mathbb{N}, \{0, 1\}, +, >)$, describes the properties of the natural numbers without the multiplication~\cite{presburger1929}, 
the Zermelo-Fraenkel set theory is the set of the formulas defining the notion of set, and Tarski's theory~\cite{Fraenkel1958}, also known as semi-algebraic theory
over the reals,
$(\mathbb{R}, \{0, 1\}, +, *, >)$, is the set of the first-order formulas whose expressions are polynomials with integer coefficients~\cite{Tarski51}.

A set $S$ is \emph{definable} in a theory $\mathcal{T}$ if there exists a formula 
$\psi(x) \in \mathcal{T}$ such that $S=\{x\, |\, \psi(x)\}$. 
\begin{example} 
Let us consider the set $S$ of the pairs $\langle x, y \rangle \in \mathbb{R}^2$ such that $1/(x*y)>1$, i.e., $S=\{\langle x, y \rangle \in \mathbb{R}^2\,|\, 1/(x*y) > 1\}$.
The formula $1/(x*y)>1$ does not belong to Tarski's theory because the left expression includes a division, which is not one of the $(\mathbb{R}, \{0, 1\}, +, *, >)$'s 
functions. 
However, the formulas $(y > 0 \land x > 0 \land 1>x*y) \lor (0 > y \land 0 > x \land 1>x*y)$ and $1/(x*y)>1$ are equivalent on $\mathbb{R}$ 
and the former belongs to $(\mathbb{R}, \{0, 1\}, +, *, >)$. Thus, the set $S$ is definable in Tarski's theory.
\end{example}
A theory $\mathcal{T}$ over $\mathcal{Q}$ is \emph{o-minimal} if any set 
$S \subseteq \mathcal{Q}$, definable  $\mathcal{T}$, in is the finite union of open intervals and points~\cite{Dries_1998}. 
Tarski's theory and its Pfaffian extensions~\cite{speissegger1999pfaffian}, e.g.,  $(\mathbb{R},  \{0,1\}, +, $ $*,e^{x},  >)$ or $(\mathbb{R},  \{0,1\}, +, $ $*,\ln{x},  >)$, 
are o-minimal theories.
An \emph{o-minimal set} is a set definable in some o-minimal theory.

\section{Significativity over Confusion Matrices}\label{sec:significativity_confusion}

Let $\sigma$ be a quality-ordered agreement measure for $n\times n$ matrices. 

The ratio between the number of confusion matrices in $\mathcal{M}_{n,m}$ to be such that $\sigma(M) < c$ 
and the total number of confusion matrices in $\mathcal{M}_{n,m}$ is
\begin{equation}\label{eq:varrho}
\varrho_{\sigma,n,m}(c) \defeq \frac{\left|\left\{M \in \mathcal{M}_{n,m}\, |\, \sigma(M) < c \right\}\right|}{\left|\mathcal{M}_{n,m}\right|}.
\end{equation}
This value belongs to the real interval $[0,1]$ and reports how many of all the $n \times n$ confusion matrices consisting of $m$ tests have 
an agreement value less than $c$. From a probabilistic point of view, $\varrho_{\sigma,n,m}(c)$ is also 
the probability of selecting by chance over a uniform distribution a $n\times n$-confusion matrix of $m$ tests whose agreement value 
is less than $c$. Because of this, we say that $\varrho_{\sigma,n,m}(c)$ is
\emph{the $\sigma$-significativity of $c$ in $\mathcal{M}_{n,m}$}.

The reader must be aware that the $\sigma$-significativity does not rate the confusion matrices, which is what the agreement measure $\sigma$ deals with.
Instead, it provides a significativity measure for the agreement values.

\begin{figure*}[!ht]
     \begin{subfigure}[b]{0.49\textwidth}
\resizebox{\textwidth}{!}{
\begin{tikzpicture}
    \begin{axis}[xmin=-1,xmax=1,
                        ymin=0,ymax=1,
                        xlabel={$c$},
                        ylabel={$\kappa$-Significativity},
      			legend pos=north west,
			legend style={cells={align=left}}]
	\addplot table [x=kappa,y=R_kappa_2_2,col sep=space] {kappa_2.csv};
	\addlegendentry{$\varrho_{\kappa,2,2}(c)$}
	\addplot table [x=kappa,y=R_kappa_2_4,col sep=space] {kappa_2.csv};
	\addlegendentry{$\varrho_{\kappa,2,4}(c)$}
	\addplot table [x=kappa,y=R_kappa_2_16,col sep=space] {kappa_2.csv};
	\addlegendentry{$\varrho_{\kappa,2,16}(c)$}
    \end{axis}
\end{tikzpicture}}
\caption{The function $\varrho_{\kappa,2,m}(c)$.}
\end{subfigure}
%\hfill
%     \begin{subfigure}[b]{0.31\textwidth}
%\resizebox{\textwidth}{!}{
%\begin{tikzpicture}
%    \begin{axis}[xmin=0,xmax=1,
%                        ymin=0,ymax=1,
%                        xlabel={$x$},
%                        ylabel={\% of $2\times 2$-matrices (Probability)},
%      			legend pos=south east,
%			legend style={cells={align=left}}]
%	\addplot table [x=ir,y=p,col sep=space] {IR_rho.csv};
%	\addlegendentry{$\rho_{\IR}(x)$}
%\end{axis}
%\end{tikzpicture}}
%\caption{The function $\rho_{\IR}(x)$.}
%\end{subfigure}
\hfill
\begin{subfigure}[b]{0.49\textwidth}
\resizebox{\textwidth}{!}{
\begin{tikzpicture}
    \begin{axis}[xmin=-1,xmax=1,
                        ymin=0,ymax=1,
                        xlabel={$c$},
                        ylabel={$\kappa$-Significativity},
      			legend pos=north west,
			legend style={cells={align=left}}]
	\addplot table [x=kappa,y=R_kappa_3_2,col sep=space] {kappa_3.csv};
	\addlegendentry{$\varrho_{\kappa,3,2}(c)$}
	\addplot table [x=kappa,y=R_kappa_3_4,col sep=space] {kappa_3.csv};
	\addlegendentry{$\varrho_{\kappa,3,4}(c)$}
	\addplot table [x=kappa,y=R_kappa_3_8,col sep=space] {kappa_3.csv};
	\addlegendentry{$\varrho_{\kappa,3,8}(c)$}
    \end{axis}
\end{tikzpicture}}
\caption{The function $\varrho_{\kappa,3,m}(c)$.}
\end{subfigure}
\caption{The functions $\varrho_{\kappa,n,m}(c)$ for $n=2$ and $n=3$ as the number of tests $m$ changes.}\label{fig:varrho-kappa}
\end{figure*}
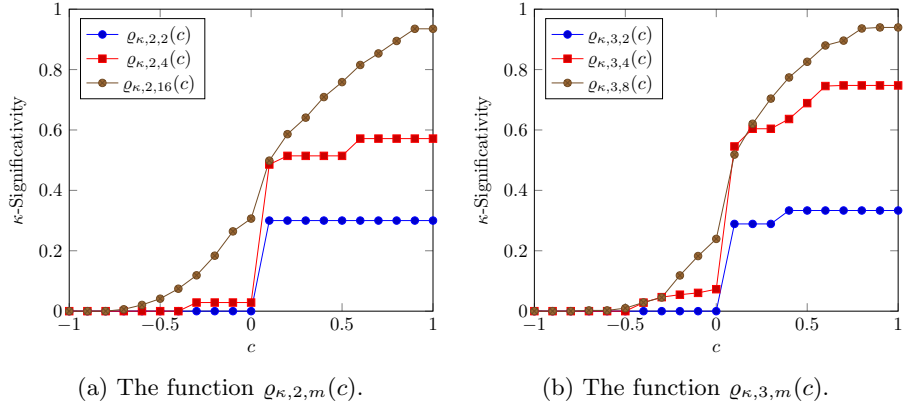

%
%\begin{algorithm}
%\caption{An iterator for the set $\mathcal{C}_{m,k}$. \textbf{Complexity $\Theta\left(k\binom{m+k-1}{m}\right)$}?}\label{alg:wc_iterator}
%\begin{algorithmic}[1]
%\Require $m \in \mathbb{N}$, $k \in \mathbb{N}_{>0}$, and $x \in \mathbb{N}^i$ for some $i \geq k$.
%When $x$ is not provided, it is initialised to  $\langle 0, \ldots, 0 \rangle\in \mathbb{N}^k$.
%\Ensure To yield all the weak compositions of $m$ in $k$ parts.
%\Procedure{WeakCompositions}{$m, k, x \gets \langle 0, \ldots, 0 \rangle$}
%\If{$k=1$}
%\State $x[k] \gets m$
%\State \textbf{yield} $x$
%\Else
%\For{$v \gets 0, \dots, m$}
%\State $x[k] \gets m$
%\For{$y \gets $\Call{WeakCompositions}{$m-v, k-1, x$}}
%\State \textbf{yield} $y$
%\EndFor
%\EndFor
%\EndIf
%\EndProcedure
%\end{algorithmic}
%\end{algorithm}

%\subsection{Computing the Significativity}~\label{sec:significativity_computation}

A \emph{weak composition of $m$ in $k$ parts} is a tuple $\langle x_1, \ldots, x_k \rangle \in \mathbb{N}^{k}$ such that 
$\sum_{i=1}^{k} x_i = m$~\cite{bona2016}. The set $\mathcal{C}_{m,k}$ that contains all of them, i.e., 
\begin{equation}\label{eq:compositions}
\mathcal{C}_{m,k} \defeq \left\{\langle x_1, \ldots, x_k \rangle \in \mathbb{N}^{k} \,|\, \sum_{i=1}^{k} x_i = m\right\}
\end{equation}
has cardinality $\binom{m+k-1}{m}$~\cite[Theorem~5.2]{bona2016}.

Let  $\gamma: \mathbb{R}^{n^2} \rightarrow \mathbb{R}^{n \times n}$ be defined as
\begin{equation}
\gamma(\langle x_1, \ldots, x_{n^2} \rangle) \defeq \begin{pmatrix}
x_1 & \ldots & x_n\\
x_{n+1} & \ldots & x_{2n}\\
\vdots & \ddots & \vdots&\\
x_{n(n^2-1)+1} & \ldots & x_{n^2} 
\end{pmatrix}.
\end{equation}
It is easy to see that $\gamma$ is bijective and maps any weak composition of $m$ in $n^2$ parts 
into a $n\times n$ confusion matrix over $m$ tests, i.e., $\mathcal{M}_{n,m} = \gamma\left(\mathcal{C}_{m,n^2}\right)$ and 
$\mathcal{C}_{m,n^2} = \gamma^{-1}\left(\mathcal{M}_{n,m}\right)$.
%and $\left|\mathcal{M}_{n,m}\right|$
%equals $\left|\mathcal{C}_{n^2,m}\right|$ which is $\binom{m+n^2-1}{m}$~\cite[Theorem~5.2]{bona2016}.
Hence, the sets $\left\{M \in \mathcal{M}_{n,m}\, |\, \sigma(M) < c \right\}$  and 
\begin{equation}
\mathcal{R}_{\sigma,n,m}(c)\defeq  \left\{x \in \mathcal{C}_{m,n^2}\, |\, \sigma(\gamma(x)) < c \right\}
\end{equation}
have the same cardinality. Moreover, $|\mathcal{M}_{n,m}|=|\mathcal{C}_{m,n^2}|$ because
$\gamma$ is bijective.
Thus, $\varrho_{\sigma,n,m}(c) = |\mathcal{R}_{\sigma,n,m}(c)|/|\mathcal{C}_{m,n^2}|$.
The cardinality of $\mathcal{C}_{m,n^2}$ is 
$\binom{m+n^2-1}{m}$~\cite[Theorem~5.2]{bona2016} 
%If $\left|\mathcal{C}_{m,n^2}\right| = \binom{m+n^2-1}{m}$ is reasonably small, 
and we can compute $\left|\mathcal{R}_{\sigma,n,m}(c)\right|$ by iterating over 
all the elements in $\mathcal{C}_{m,n^2}$ %(for instance, by using Algorithm~\ref{alg:wc_iterator})
and summing up their images through the indicator function of $\mathcal{R}_{\sigma,n,m}(c)$, i.e., 
$\left|\mathcal{R}_{\sigma,n,m}(c)\right|= \sum_{x \in \mathcal{C}_{m,n^2}} \mathbf{1}_{\mathcal{R}_{\sigma,n,m}(c)}(x)$ 
(see Algorithm~\ref{alg:R_computation}).
%This is the case in clinical studies for dichotomous rare disease tests involving up to a few tens of tests, i.e., $n=2$ and $m<50$.

\begin{algorithm}
\caption{An algorithm to compute $|\mathcal{R}_{\sigma,n,m}(c)|$ in time $T(\sigma \circ \gamma)*\Theta\left(\left|\mathcal{C}_{m, n^2}\right|\right)$.
%The logarithmic term %$\Theta\left(\log_2{\binom{m+k-1}{m}}\right)$ 
%is due to the logarithmic cost criterion.
}\label{alg:R_computation}
\small
\begin{algorithmic}[1]
\Require $m \in \mathbb{N}$, $n \in \mathbb{N}_{>0}$, $c \in \mathbb{R}$, and $\sigma: \mathcal{M}_{n,m} \rightarrow \mathbb{R}$.
\Ensure Returns $|\mathcal{R}_{\sigma,n,m}(c)|$.
\Function{aux\_C}{$\sigma, m, k, c, x$}\Comment{An auxiliary function}
\If{$k=1$}
\State $x[k] \gets m$\Comment{$x$ is now belong to $\mathcal{C}_{m, n^2}$}
\State \Return $(1 \textbf{ if } \sigma(\gamma(x))< c \textbf{ else } 0)$%\Comment{Evaluate $\mathbf{1}_{\mathcal{R}_{\sigma,n,m}(c)}(\gamma(x))$}
\EndIf
\State $c \gets 0$%\Comment{$c$ is a counter for the elements in $\mathcal{R}_{\sigma,n,m}(c)$}
\For{$v \gets 0, \dots, m$}
\State $x[k] \gets v$
\State $c \gets counter+$\Call{aux\_C}{$\sigma, m-v, k-1, c, x$} %\Comment{This sum costs $\Theta\left(\log_2{\binom{m+k-1}{m}}\right)$ according to the logarithmic cost criterion}
\EndFor
\State \Return $c$
\EndFunction
\Statex
\Function{R\_Cardinality}{$\sigma, n, m, c$}
\State $x \gets $\Call{array}{$n^2, 0$} \Comment{$x$ is initialized to $\vec{0}\in \mathbb{N}^{n^2}$}
\State \Call{aux\_C}{$\sigma, m, n^2, c, x$}
\EndFunction
\end{algorithmic}
\end{algorithm}

The asymptotic time cost~\cite{cormen2022introduction} of evaluating the function $\gamma$ is $O(n^2)$. 
Thus, the complexity of the presented 
approach to compute $\varrho_{\sigma,n,m}(c)$ is  $(T(\sigma) + O(n^2))*\Theta(\binom{m+n^2-1}{m})$, where 
$T(f)$ is the time complexity of the function $f$.
It follows that the complexity of the approach is $\Theta(n^2 \binom{m+n^2-1}{m})$ when 
$T(\sigma) \in \Theta(n^2)$ as in the cases of Cohen's $\kappa$ or $\IA$.
%
%\begin{example}
%Let $M_C$ be the matrix  
%\[
%M_C=\begin{pmatrix}\num{19804}&\num{172}\\\num{1}&\num{23}\end{pmatrix}.
%\] 
%Cohen's $\kappa$ and $\IA$ for $M_C$ are $\kappa(M_C)\approx 0.20835$ and $\IA(M_C)\approx 0.55963$, respectively.
%The matrix $M_C$ summarises the results of $m=\num{19804}+\num{172}+\num{1}+\num{23}=\num{20000}$ tests. 
%The $\kappa$-significativity of $\kappa(M_C)$ in $\mathcal{M}_{2,\num{20000}}$ is $\varrho_{\kappa,2,\num{20000}}(\kappa(M_C))\approx 0.7259$. 
%Thus, less than $73\%$ of the $2\times 2$-confusion matrices over $\num{20000}$ tests have a $\kappa$ value smaller than that of $M_C$ and 
%the probability of choosing by chance a confusion matrix $M_C' \in \mathcal{M}_{2,\num{20000}}$ with $\kappa(M_C') \leq \kappa(M_C)$ is 
%$0.7259$.
%
%Instead, the $\IA$-significativity of $\IA(M_C)$ in $\mathcal{M}_{2,\num{20000}}$ is $\varrho_{\IA,2,20}(\IA(M_C))\approx 0.9611$.
%Hence, more than $96\%$ of the  $2\times 2$-confusion matrices over $\num{20000}$ tests have an $\IA$ value smaller than that of $M_C$  and 
%the probability of choosing by chance a confusion matrix $M_C' \in \mathcal{M}_{2,\num{20000}}$ with $\IA(M_C') \leq \IA(M_C)$ is 
%$0.9611$.
%\end{example}

\begin{example}\label{ex:varrho}
Let $M_C$ be the matrix  
\[
M_C=\begin{pmatrix}\num{8}&\num{3}\\\num{0}&\num{9}\end{pmatrix}.
\] 
Cohen's $\kappa$ and $\IA$ for $M_C$ are $\kappa(M_C)\approx 0.70588$ and $\IA(M_C)\approx 0.52115$, respectively.
The matrix $M_C$ summarises the results of $m=\num{8}+\num{3}+\num{0}+\num{9}=\num{20}$ tests. 
The $\kappa$-significativity of $\kappa(M_C)$ in $\mathcal{M}_{2,\num{20}}$ is $\varrho_{\kappa,2,\num{20}}(\kappa(M_C))\approx 0.8866$. 
Thus, more than $88\%$ of the $2\times 2$-confusion matrices over $\num{20}$ tests have a $\kappa$ value lower than that of $M_C$ and 
the probability of choosing by chance a confusion matrix $M_C' \in \mathcal{M}_{2,\num{20}}$ with $\kappa(M_C') < \kappa(M_C)$ is 
$0.8866$.

Instead, the $\IA$-significativity of $\IA(M_C)$ in $\mathcal{M}_{2,\num{20}}$ is $\varrho_{\IA,2,20}(\IA(M_C))\approx 0.7628$.
Hence, more than $76\%$ of the  $2\times 2$-confusion matrices over $\num{20}$ tests have an $\IA$ value lower than that of $M_C$. Moreover,
the probability of choosing by chance a confusion matrix $M_C' \in \mathcal{M}_{2,\num{20}}$ with $\IA(M_C') < \IA(M_C)$ is 
$0.2324$.
\end{example}

Figure~\ref{fig:varrho-kappa} clarifies the relation between $n$, $m$, and $\varrho_{\kappa, n, m}(c)$ 
for $n$ in ${2,3}$ as $m$ grows.

It is worth remarking that Cohen's $\kappa$, as many other agreement measures, assumes values in $[-1,1]$ where the negative part
of the interval is usually associated with confusion matrices exhibiting ``disagreement'' among
the classifiers. The $\kappa$-significativity does not distinguish between agreement and ``disagreement''.
Hence, when, in Example~\ref{ex:varrho}, we wrote that more than $95\%$ of the matrices in
$\mathcal{M}_{2,\num{20}}$ have a $\kappa$ value smaller than that of $M_C$, we referred to the full
set of matrices in $\mathcal{M}_{2,\num{20}}$ with no reference to sign of their $\kappa$-images.

\subsection{Numerical Evaluation}

Due of its time complexity, the approach sketched in Section~\ref{sec:significativity_confusion} fails to scale up to trials dealing 
with hundreds of tests or involving non-dichotomic outcomes. For instance, an evaluation of the probability for a $2\times 2$
confusion matrix $M$ summarising $200$ test results that $\sigma(M) < c$, i.e., $\varrho_{\sigma,2,200}(c)$, iterates over 
$\binom{200+2^2-1}{200} = \num{1373701}$ weak compositions, which are less than half of the $\binom{20+3^2-1}{20} = \num{3108105}$ weak compositions
required to evaluate the probability for a $3\times 3$
confusion matrix $M$ collecting $20$ test results of satisfying the same inequality, i.e., $\varrho_{\sigma,3,20}(c)$.

%For instance, $\iota_{k,m}$ can enumerate
%all the weak compositions of $m$ into $k$ in lexicographic order, i.e., if $a,b \in \left[0, \binom{m+k-1}{m}-1\right]$ and $a<b$, 
%then there exists $j \in [1, k]$ such that 
%$a_i = b_i$ for all $i \in [1,j-1]$ and $a_j < b_j$ where $\iota_{k,m}'(a)=\langle a_1, \ldots, a_k \rangle$ and 
%$\iota_{k,m}'(b)=\langle b_1, \ldots, b_k \rangle$.

Nevertheless, $\varrho_{\sigma,n,m}(c)$ can be numerically estimated by using the Monte Carlo method~\cite{montecarlo} as 
\begin{equation}\label{eq:monte_carlo_varrho}
\varrho_{\sigma,n,m}(c) \approx 
%\frac{\frac{\left|\mathcal{C}_{m,n^2}\right|}{N} \sum_{i=1}^{N} \mathbf{1}_{\mathcal{R}_{\sigma,n,m}(c)}(x_i)}{\left|\mathcal{C}_{m,n^2}\right|} = 
\frac{1}{N} \sum_{i=1}^{N} \mathbf{1}_{\mathcal{R}_{\sigma,n,m}(c)}(x_i)
\end{equation}
where $x_1, \ldots, x_N$ are $N$ weak compositions uniformly distributed in $\mathcal{C}_{m,n^2}$.

Since $\mathcal{C}_{m,k}$ is discrete, there is a bijective function 
%$\iota_{m,k}: \left[0, \binom{m+k-1}{m}-1\right] \rightarrow \mathcal{C}_{m,k}$
$\iota_{m,k}: \left[0, \left|\mathcal{C}_{m,k}\right|-1\right] \rightarrow \mathcal{C}_{m,k}$ 
mapping each natural number lower than 
$\left|\mathcal{C}_{m,k}\right|$ to a weak composition of $m$ into $k$ parts  -- actually, there 
are $\left|\mathcal{C}_{m,k}\right|!$ of them because $\mathcal{C}_{m,k}$ is finite --.
If we apply $\iota_{m,n^2}$ to uniform samples over the integers in $\left[0, \left|\mathcal{C}_{m,n^2}\right|-1\right]$, 
we get uniform samples over $\mathcal{C}_{m,n^2}$, and we can approximate $\varrho_{\sigma,n,m}(c)$ as suggested by Eq.~\ref{eq:monte_carlo_varrho} 
with an error proportional to $1/\sqrt{N}$ (e.g., see~{\cite{Mackay1998}).
Figure~\ref{fig:montecarlo_error} represents the average error of 100 Monte Carlo-based estimations
of $\varrho_{\kappa, n, m}(c)$ using 
$\left\lceil\sqrt{\left|\mathcal{C}_{m,n^2}\right|}\right\rceil$ samples.

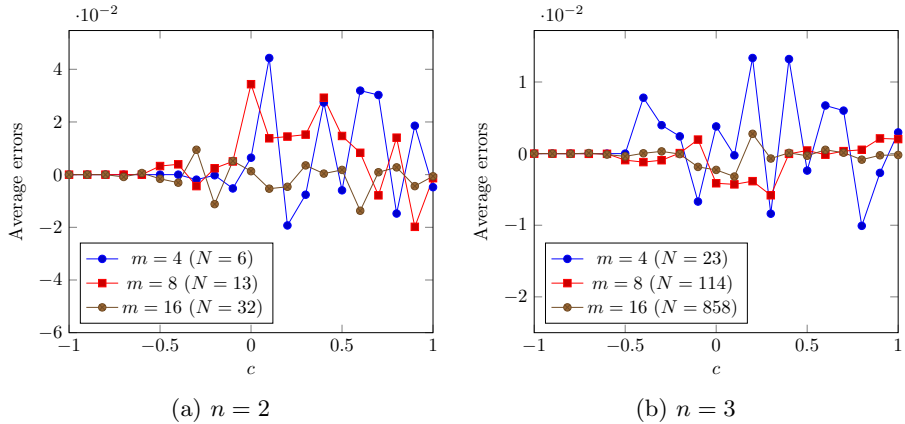
\begin{figure*}[!ht]
     \begin{subfigure}[b]{0.49\textwidth}
\resizebox{\textwidth}{!}{
\begin{tikzpicture}
    \begin{axis}[xmin=-1,xmax=1,
    		       ymin=-6e-2,
                        xlabel={$c$},
                        ylabel={Average errors},
      			legend pos=south west,
			legend style={cells={align=left}}]
	%\addplot table [x=kappa,y=mean_delta_R_kappa_2_2,col sep=space] {kappa_2.csv};
	%\addlegendentry{$m=2$}
	\addplot table [x=kappa,y=mean_delta_R_kappa_2_4,col sep=space] {kappa_2.csv};
	\addlegendentry{$m=4$ $(N=6)$}
	\addplot table [x=kappa,y=mean_delta_R_kappa_2_8,col sep=space] {kappa_2.csv};
	\addlegendentry{$m=8$  $(N=13)$}
	\addplot table [x=kappa,y=mean_delta_R_kappa_2_16,col sep=space] {kappa_2.csv};
	\addlegendentry{$m=16$  $(N=32)$}
    \end{axis}
\end{tikzpicture}}
\caption{$n=2$}
\end{subfigure}
%\hfill
%     \begin{subfigure}[b]{0.31\textwidth}
%\resizebox{\textwidth}{!}{
%\begin{tikzpicture}
%    \begin{axis}[xmin=0,xmax=1,
%                        ymin=0,ymax=1,
%                        xlabel={$x$},
%                        ylabel={\% of $2\times 2$-matrices (Probability)},
%      			legend pos=south east,
%			legend style={cells={align=left}}]
%	\addplot table [x=ir,y=p,col sep=space] {IR_rho.csv};
%	\addlegendentry{$\rho_{\IR}(x)$}
%\end{axis}
%\end{tikzpicture}}
%\caption{The function $\rho_{\IR}(x)$.}
%\end{subfigure}
\hfill
\begin{subfigure}[b]{0.49\textwidth}
\resizebox{\textwidth}{!}{
\begin{tikzpicture}
    \begin{axis}[xmin=-1,xmax=1,
    		       ymin=-3.5e-2,
                        xlabel={$c$},
                        ylabel={Average errors},
      			legend pos=south west,
			legend style={cells={align=left}}]
	%\addplot table [x=kappa,y=mean_delta_R_kappa_3_2,col sep=space] {kappa_3.csv};
	%\addlegendentry{$m=2$}
	\addplot table [x=kappa,y=mean_delta_R_kappa_3_4,col sep=space] {kappa_3.csv};
	\addlegendentry{$m=4$ $(N=23)$}
	\addplot table [x=kappa,y=mean_delta_R_kappa_3_8,col sep=space] {kappa_3.csv};
	\addlegendentry{$m=8$ $(N=114)$}
	\addplot table [x=kappa,y=mean_delta_R_kappa_3_16,col sep=space] {kappa_3.csv};
	\addlegendentry{$m=16$ $(N=858)$}
    \end{axis}
\end{tikzpicture}}
\caption{$n=3$}
\end{subfigure}
\caption{The average difference between $\varrho_{\kappa,n,m}(c)$ and $100$ of its Monte Carlo estimation when 
$N= \left\lceil\sqrt{\left|\mathcal{C}_{m,n^2}\right|}\right\rceil$.}\label{fig:montecarlo_error}
\end{figure*}

In this case, the time complexity is $N *  (T(\sigma) +T(\gamma) + T(\iota_{m, n^2}) + T(\mathcal{N}_{n,m})) $ where $\mathcal{N}_{n,m}$ 
is the function that uniformly samples the natural numbers in $\left[0, \left|\mathcal{C}_{m,n^2}\right|-1\right]$. 
The most used pseudo-random number generators, such as Mersenne Twister~\cite{10.1145/272991.272995}, WELL~\cite{10.1145/1132973.1132974},   
and \texttt{xoroshiro128++}~\cite{10.1145/3460772}, takes linear time on the number of output bits per generation, i.e., $\Theta\left(\log_2{\left|\mathcal{C}_{m,n^2}\right|}\right)$.
However, if %we assume to exclusively deal with $n$s and $m$s such that $\log_2{\left|\mathcal{C}_{m,n^2}\right|} < c_\text{bits}$  for some 
%constant $c_\text{bits}$. 
we assume  the number of bits in the representation of both $n$ and $m$ to be upper-bound, then $T(\mathcal{N}_{n,m}) \in \Theta(1)$.
This constrain may be reasonable in the investigated domain where the matrix size, $n$, and the number of 
tests used to build the confusion matrices, $m$, are usually upper bounded by $5$ 
(e.g., PI-RADS~\cite{PIRADS2}, BI-RADS~\cite{birads2013}) and $\num{1000000}$ (for AI applications), respectively.
Under these conditions, the Monte Carlo method takes time  
$N *  (T(\sigma) + T(\iota_{m, n^2}))$ to evaluate $\varrho_{\sigma,n,m}(c)$ 
using $N$ samples. If $T(\sigma) \in \Theta(n^2)$, as in the case of Cohen's $\kappa$, Fleiss's $\kappa$, $\IA$, and $\IR$, the time complexity belongs to  
$N *  (T(\iota_{m, n^2}) + \Theta(n^2))$. 

When the bits in the number representation cannot be upper-bounded, a more refined analysis is required to access the asymptotic complexity of the proposed method. In particular, the logarithmic cost criterion~\cite{cormen2022introduction} should be used to take into account that the complexity of each arithmetic operation is affected by the sizes of the operator representations. 

The following section presents the lexicographic order enumerator for the set $\mathcal{C}_{m,k}$ and 
proves that $T(\iota_{m, k}) \in O(k+m)$. 

%Thus, when $\iota_{m, n^2}$ is the lexicographic order enumerator for $\mathcal{C}_{m,n^2}$, 
%the estimating   $\varrho_{\sigma,n,m}(c)$ by Monte Carlo method takes time $O(N(n^2+m))$.

\subsection{The Lexicographic Order Enumeration of $\mathcal{C}_{m,k}$}

The relation $<_l \subseteq \mathcal{C}_{m,k} \times \mathcal{C}_{m,k}$ is the \emph{lexicographic order} among the weak compositions in $\mathcal{C}_{m,k}$
when for any $a, b \in\mathcal{C}_{m,k}$, $a <_l b$ implies that there exists $j \in [1, k]$ such that $a[i] = b[i]$ for all $i \in [1,j-1]$ and $a[j] < b[j]$.

The \emph{lexicographic order enumerator of set $\mathcal{C}_{m,k}$} is
the bijective function $\iota_{m,k}: \left[0, \left|\mathcal{C}_{m,k}\right|-1\right] \rightarrow \mathcal{C}_{m,k}$ such that $i<j$ if and only if $\iota_{m,k}(i) <_l  \iota_{m,k}(j)$ for any $i,j \in \left[0, \left|\mathcal{C}_{m,k}\right|-1\right]$.
Thus, $\iota_{m,k}(i)=\langle\ell_1, \ldots, \ell_k\rangle$ is the $i$-th element in the lexicographic order among the weak compositions in $\mathcal{C}_{m,k}$.
This section presents a time-efficient algorithm to evaluate $\iota_{m,k}$. 

Let us consider the weak compositions in $\mathcal{C}_{m,k}$ having $j \in [0, m]$ as the first component. 
Since they belong to $\mathcal{C}_{m,k}$, 
the sum of their components is $m$. Thus, there are as many as the weak compositions of $m-j$ in $k-1$ parts because their first component is $j$, i.e., there are 
$\left|\mathcal{C}_{m-j,k-1}\right|$ weak compositions in $\mathcal{C}_{m,k}$ whose first component is $j$.

Because of the definition of lexicographic order, the first weak compositions in the lexicographic order are those having $0$ as the first component. 
Then there are those having $1$ as
the first component, and so on up to the weak compositions whose first component is $m$. It follows that $i\geq\left|\mathcal{C}_{m,k-1}\right|$ if and only if 
$\iota_{m,k}(i)$ follows all the weak compositions whose first component is $0$ in lexicographic order.
Analogously,  $i\geq\left|\mathcal{C}_{m,k-1}\right|+\left|\mathcal{C}_{m-1,k-1}\right|$ if and only if $\iota_{m,k}(i)$ follows 
all the weak compositions whose first component is $1$  in the lexicographic order. In general, $i\geq \sum_{j=0}^{l}\left|\mathcal{C}_{m-j,k-1}\right|$ if and only if 
$\iota_{m,k}(i)$ follows all the weak compositions whose first component is $l$. As a consequence, the first component of $\iota_{m,k}(i)$, i.e., $\ell_1$, is such that 
\begin{equation}
\sum_{j=0}^{\ell_1-1}\left|\mathcal{C}_{m-j,k-1}\right| \leq i < \left|\mathcal{C}_{m-\ell_1,k-1}\right| + \sum_{j=0}^{\ell_1-1}\left|\mathcal{C}_{m-j,k-1}\right|
\end{equation} 
or, equivalently,
\begin{equation}\label{eq:i_succession}
0\leq i -\sum_{j=0}^{\ell_1-1}\left|\mathcal{C}_{m-j,k-1}\right| < \left|\mathcal{C}_{m-\ell_1,k-1}\right|.
\end{equation} 

The following theorem suggests a strategy to identify the remaining components of $\iota_{m,k}(i)$, i.e., $\ell_2, \ldots, \ell_k$.
\begin{theorem}\label{theorem:iota1}
For any $m \in \mathbb{N}$, for any $k \in \mathbb{N}$, and for any $i \in \mathbb{N}$, if $k>1$ and $\iota_{m,k}(i)=\langle \ell_1, \ell_2, \ldots, \ell_{k}\rangle$, then 
\begin{equation}
\iota_{m', k-1}(i')=\langle \ell_2, \ldots, \ell_{k} \rangle
\end{equation} 
where $m' = m - \ell_1$ and $i' = i -\sum_{j=0}^{\ell_1-1}\left|\mathcal{C}_{m-j,k-1}\right|$.
\end{theorem}
\begin{proof}
Let $<_l'$ be the lexicographic order among weak compositions in $\mathcal{C}_{m-\ell_1,k-1}$. According to the definition of lexicographic order, 
$\langle a_1, \ldots, a_{k-1} \rangle <_l' \langle b_1, \ldots, b_{k-1} \rangle$ if and only if there exists $j \in [1, k-1]$ such that for all $i \in [1, j-1]$ 
$a_i = b_i$ and $a_j < b_j$. Thus, $\langle a_1, \ldots, a_{k-1} \rangle <_l' \langle b_1, \ldots, b_{k-1} \rangle$ if and only if 
$\langle \ell_1, a_1, \ldots, a_{k-1} \rangle <_l' \langle \ell_1, b_1, \ldots, b_{k-1} \rangle$ and the lexicographic order among the weak compositions of $m$
in $k$ parts whose first component is $\ell_1$ and $<_l'$ are consistent.
Hence, the components $\ell_2', \dots, \ell_k'$ of $i'$-th weak composition with respect to the $<_l'$
 must be the $i'$-th weak composition having $\ell_1$ as the first component with respect to the $<_l$, i.e., 
$\iota_{m-\ell_1, k-1}(i')=\langle \ell_2', \ldots, \ell_{k}' \rangle$ if and only if $\langle \ell_1, \ell_2', \ldots, \ell_{k}' \rangle$ is the 
 $i'$-th weak composition having $\ell_1$ as the first component with respect to the $<_l$.
Because of the definition of lexicographic order, 
for all $a, b \in \mathcal{C}_{m,k}$, if the first components of $a$ and $b$ are $a_1$ and $b_1$ and $a_1 \neq b_1$, then 
 $a <_l b$ if and only if $a_1 < b_1$. Hence, all the weak compositions in $\mathcal{C}_{m,k}$ whose first component is lower than 
 $\ell_1$ come before the weak compositions in $\mathcal{C}_{m,k}$ whose first component is $\ell_1$ according to $<_l$. 
 Analogously, all the weak compositions in $\mathcal{C}_{m,k}$ whose first component is greater than 
 $\ell_1$ come after the weak compositions in $\mathcal{C}_{m,k}$ whose first component is $\ell_1$ according to $<_l$. 
There are $\left|\mathcal{C}_{m-j,k-1}\right|$ weak compositions in $\mathcal{C}_{m,k}$ whose first component is $j$.
Thus, the $i'$-th weak composition in $\mathcal{C}_{m,k}$ that has $\ell_1$ as the first component with respect to the $<_l$ 
is in position $i'+\sum_{j=0}^{\ell_1-1}\left|\mathcal{C}_{m-j,k-1}\right|$ of the overall lexicographic order among all 
the weak compositions in $\mathcal{C}_{m,k}$. Since $i'=i-\sum_{j=0}^{\ell_1-1}\left|\mathcal{C}_{m-j,k-1}\right|$ by hypothesis, 
$i'+\sum_{j=0}^{\ell_1-1}\left|\mathcal{C}_{m-j,k-1}\right|=(i-\sum_{j=0}^{\ell_1-1}\left|\mathcal{C}_{m-j,k-1}\right|)+\sum_{j=0}^{\ell_1-1}\left|\mathcal{C}_{m-j,k-1}\right|=i$ 
and 
the $i'$-th weak composition in $\mathcal{C}_{m,k}$ that has $\ell_1$ as the first component with respect to the $<_l$ 
is in position $i$ of the overall lexicographic order among all 
the weak compositions in $\mathcal{C}_{m,k}$. It follows that 
$\iota_{m-\ell_1, k-1}(i')=\langle \ell_2', \ldots, \ell_{k}' \rangle$ if and only if $\langle \ell_1, \ell_2', \ldots, \ell_{k}' \rangle$ is the 
$i$-th weak composition having $\ell_1$ as the first component with respect to the $<_l$, i.e., 
$\iota_{m-\ell_1, k-1}(i')=\langle \ell_2', \ldots, \ell_{k}' \rangle$ if and only if $\iota_{m, k}(i)=\langle \ell_1, \ell_2', \ldots, \ell_{k}' \rangle$.
\end{proof}

By exploiting Theorem~\ref{theorem:iota1}, we can build an algorithm that iteratively identifies all the components of $\iota_{m,k}(i)$ 
in the order of the components themselves.
The central expression's index $j$ in Eq.~\ref{eq:i_succession} ranges from $0$ up to $\ell_1-1$. The same expression can be re-written 
by using an index from $m$ down to $m-\ell_1+1$ 
as follows
\begin{equation}
 i -\sum_{j=0}^{\ell_1-1}\left|\mathcal{C}_{m-j,k-1}\right| = i -\sum_{j=m-\ell_1+1}^{m}\left|\mathcal{C}_{j,k-1}\right|.
\end{equation} 

Hence, we can use a variable $m_l$ to store the initial value of $m$ and
decrease $i$ and $m$ by $\left|\mathcal{C}_{m,k-1}\right|$ and $1$, respectively, as long as $m>0$ and 
$i\geq \left|\mathcal{C}_{m,k-1}\right|$. 
The first component of $\iota_{m,k}(i)$ will be the difference among $m_l$ and the first $m$ such that $i < \left|\mathcal{C}_{m,k-1}\right|$, 
i.e., $\ell_1=m_l-m$. The following components can be identified by decreasing $k$ by $1$ and repeating the previous steps.

\begin{algorithm}
\caption{A lexicographic order enumerator for the set $\mathcal{C}_{m,k}$. 
The time complexity of this algorithm is $O(mk)$.
}\label{alg:wc_enumerator1}
\small
\begin{algorithmic}[1]
\Require $m \in \mathbb{N}$, $k\in \mathbb{N}_{>0}$, and $i \in [0, \left|\mathcal{C}_{m, k}\right|-1]$.
\Ensure Returns the $i$-th element in the lexicographic order of the weak compositions of $m$ in $k$ parts.
\Function{LexicographicOrder}{$m, k, i$}
\State $\ell \gets $\Call{array}{$k, 0$}\Comment{$\ell$ is $\vec{0}\in \mathbb{N}^{k}$}
\While{$k>1$}
\State $m_l \gets m$
\While{$m>0 \land i \geq \binom{m+k-2}{m}$}\label{line:while}
\State $i \gets i - \binom{m+k-2}{m}$\label{line:i_decrease}
\State $m \gets m -1$
\EndWhile
\State $\ell[\ell.size()-k] \gets m_l - m$
\State $k \gets k - 1$
\EndWhile
\State $\ell[\ell.size()-k] \gets m$
\State \Return $\ell$
\EndFunction
\end{algorithmic}
\end{algorithm}

Algorithm~\ref{alg:wc_enumerator1} describes the proposed algorithm. 
The outer while-loop can be executed $k$ times at most
because each iteration decreases $k$ by one, $k$ never increases, and the loop ends when $k\leq 1$.
Analogously, the most nested loop is repeated at most $m+k$ times in total 
because all iterations except the last one decrease $m$ by one, $m$ never increases, 
and it ends its iterations when $m \leq 0$.
The lines~\ref{line:while} and \ref{line:i_decrease} of Algorithm~\ref{alg:wc_enumerator1} require 
the evaluation of the binomial coefficient $\binom{m+k-2}{m}$. 
Since computing $\binom{a}{b}$ requires $\min\{a-b, b\}$ multiplications and 
divisions, the time complexity of Algorithm~\ref{alg:wc_enumerator1} belongs to $O\left(k\min\{k,m\}+(m+k)\min\{k,m\}\right)=O(mk)$
according to the uniform cost criterion~\cite{cormen2022introduction}.

To improve complexity, we can observe that $\left|\mathcal{C}_{m, k}\right|$, $\left|\mathcal{C}_{m-1, k}\right|$, and $\left|\mathcal{C}_{m, k-1}\right|$ are related, and once $\left|\mathcal{C}_{m, k}\right|$ has been computed, we can evaluate both $\left|\mathcal{C}_{m-1, k}\right|$ and $\left|\mathcal{C}_{m, k-1}\right|$ in constant time according to the uniform cost criterion.
\begin{lemma}\label{lemma:binom_relation}
It holds that $\left|\mathcal{C}_{m-1, k}\right| = \frac{m}{m+k-1} \left|\mathcal{C}_{m, k}\right|$, and 
$\left|\mathcal{C}_{m, k-1}\right| = \frac{k-1}{m+k-1} \left|\mathcal{C}_{m, k}\right|$ 
for any $m, k \in \mathbb{N}_{>0}$.
\end{lemma}
\begin{proof}
As far $\left|\mathcal{C}_{m-1, k}\right|$ concerns, 
\begin{align*}
\left|\mathcal{C}_{m-1, k}\right| &= \binom{(m-1)+k-1}{m-1} = \frac{(m+k-2)!}{(m-1)!(k-1)!}\\
&= \frac{m}{m+k-1}\frac{m+k-1}{m}\frac{(m+k-2)!}{(m-1)!(k-1)!}\\
&= \frac{m}{m+k-1}\frac{(m+k-1)!}{m!(k-1)!}= \frac{m}{m+k-1}\left|\mathcal{C}_{m-1, k}\right|.
\end{align*}
Analogously,
\begin{align*}
\left|\mathcal{C}_{m, k-1}\right| &= \binom{m+(k-1)-1}{m} = \frac{(m+k-2)!}{m!(k-2)!}\\
&= \frac{k-1}{m+k-1}\frac{m+k-1}{k-1}\frac{(m+k-2)!}{m!(k-2)!}\\
&= \frac{k-1}{m+k-1}\frac{(m+k-1)!}{m!(k-1)!} = \frac{k-1}{m+k-1}\left|\mathcal{C}_{m, k-1}\right|.
\end{align*}
\end{proof}

Algorithm~\ref{alg:wc_enumerator2} mimes the steps of Algorithm~\ref{alg:wc_enumerator1} 
but, thanks to Lemma~\ref{lemma:binom_relation}, avoids the computation of the binomial coefficient at each while-loop iteration.
The time complexity of Algorithm~\ref{alg:wc_enumerator2} according to the uniform const criterion 
belongs to $O(\min\{k,m\} + k + m)=O(k + m)$.

\begin{algorithm}
\caption{A lexicographic order enumerator for the set $\mathcal{C}_{m,k}$ 
%This is the fast version of the Algorithm~\ref{alg:wc_enumerator1}. It 
that avoids the computation of binomial coefficients during the while-loop iterations.
The time complexity of this algorithm is $O(m+k)$.}\label{alg:wc_enumerator2}
\small
\begin{algorithmic}[1]
\Require $m \in \mathbb{N}$, $k\in \mathbb{N}_{>0}$, and $i \in [0, \left|\mathcal{C}_{m, k}\right|-1]$.
\Ensure Returns the $i$-th element in the lexicographic order of the weak compositions of $m$ in $k$ parts.
\Function{FastLexicographicOrder}{$m, k, i$}
\State $\ell \gets $\Call{array}{$k, 0$}\Comment{$\ell$ is $\vec{0}\in \mathbb{N}^{k}$}
\State $C \gets \binom{m+k-1}{m}$\Comment{$C=\left|\mathcal{C}_{m,k}\right|$}
\While{$k>1$}
\State $C \gets \frac{k-1}{m+k-1} C$\Comment{$C=\left|\mathcal{C}_{m,k-1}\right|$}
\State $m_l \gets m$
\While{$m>0 \land i >= C$}
\State $i \gets i - C$
\State $C \gets \frac{m}{m+k-2} C$\Comment{$C=\left|\mathcal{C}_{m-1,k-1}\right|$}
\State $m \gets m-1$\Comment{$C=\left|\mathcal{C}_{m,k-1}\right|$}
\EndWhile
\State $\ell[\ell.size()-k] \gets m_l - m$
\State $k \gets k - 1$\Comment{$C=\left|\mathcal{C}_{m,k}\right|$}
\EndWhile
\State $\ell[\ell.size()-k] \gets m$
\State \Return $\ell$
\EndFunction
\end{algorithmic}
\end{algorithm}

%By exploiting Lemma~\ref{lemma:iota1}, we can define three successions of numbers $i_\alpha$, $m_\alpha$, and $\ell_\alpha$ as follows.
%
%\begin{equation}
%i_\alpha \defeq \begin{cases}
%i & \text{if $\alpha=0$}\\
%i_{\alpha-1}-\sum_{j=0}^{\ell_\alpha-1}\left|\mathcal{C}_{m_{\alpha-1}-j,k-\alpha}\right|& \text{if $\alpha>0$}
%\end{cases}
%\end{equation}
%
%\begin{equation}
%m_\alpha \defeq \begin{cases}
%m & \text{if $\alpha=0$}\\
%m_{\alpha-1}-\ell_\alpha & \text{if $\alpha>0$}
%\end{cases}
%\end{equation}
%
%\begin{equation}
%\ell_\alpha \defeq \max\left\{ \ell \in \mathbb{N} \,|\, i_\alpha > \sum_{j=0}^{\ell-1}\left|\mathcal{C}_{m_{\alpha-1}-j,k-\alpha}\right| \right\}
%\end{equation}

When $T(\sigma) \in \Theta(n^2)$ and 
$\iota_{m, n^2}$ is implemented by Algorithm~\ref{alg:wc_enumerator2},
the Monte Carlo method with $N$ samples evaluates
$\varrho_{\sigma,n,m}(c)$ in time $O(N (n^2+m))$.

\section{Significativity over Probability Matrices}\label{sec:significativity_probability}

%To compute the $\sigma$-significativity of $c$ in $\mathcal{M}_{n,m}$ we need to represent the number $\left|\mathcal{C}_{m,n^2}\right|$ 
%and this may require more bits than those available in the registers of the most common consumer 
%hardware (i.e., $64$ bit). For instance, when $n=3$ and $m=1000$, $\left|\mathcal{C}_{m,n^2}\right|$ requires
%$\lceil\log_2\left(\left|\mathcal{C}_{m,n^2}\right|\right)\rceil = 65$ bits. The same number of bits 
%are required when $n=5$ and $m=51$. Even though the asymptotic time complexity is not affected by $\mathcal{C}_{m,n^2}$ cardinality 
%when we assume $\left|\mathcal{C}_{m,n^2}\right|$ to be upper bounded by a constant, there are computational drawbacks 
%in using software arithmetic in place of the hardware arithmetic because the hardware registers cannot represent the involved numbers.
%
%Moreover, the 
The $\sigma$-significativity of $c$ in $\mathcal{M}_{n,m}$ depends on the number $m$ of the tests accounted by the matrices 
in $\mathcal{M}_{n,m}$.
In this section, we introduce a different measure of significativity 
%whose goal is 
%to gauge the $\sigma$-significativity of $c$ in $\mathcal{M}_{n,m}$ as $m$ tends to infinity.
%As a consequence, this new measure can estimate the maximum 
%theoretical $\sigma$-significativity in $\mathcal{M}_{n,m}$ for any $\sigma$-value $c$. 
for agreement values to avoid dependency on the data set size. With this aim, we
focus on probability matrices in place of confusion matrices.

The \emph{$\sigma$-significativity of $c$ in $\mathcal{P}_n$} is the ratio between the number of probability 
matrices $M_P \in \mathcal{P}_n$ such that $\sigma(M_P)< c$ and the overall number of matrices in $\mathcal{P}_n$.
These two numbers are infinite, but if $\mathcal{P}_n$ and the set of the probability matrices $M_P \in \mathcal{P}_n$ such that  $\sigma(M_P) < c$ 
are Lebesgue-measurable (e.g., see~\cite{Apostol:105425}) in an opportune space and non-null, we can evaluate the ratio between their cardinalities as the ratio 
between their Lebesgue measures.

% To evaluate this ratio, we first need a way to assess the number of probability matrices. Of course, the matrices are infinite, but we can estimate it as a ratio.
%Dealing with tests having $n$ possible outcomes, the size of the probability matrices is $n \times n$. Thus, 
According to the definition of probability matrix, a $n \times n$ probability matrix consists of $n^2$ values in the interval $[0,1]$ such that their sum 
equals $1$. Thus, we can map any $n \times n$ probability matrix into a point of the $(n^2)$-dimensional hypercube $[0,1]^{n^2}$ by using $\gamma^{-1}$.
However, not all the points in $[0,1]^{n^2}$ correspond to a probability matrix because their components must always sum up to $1$.
%Moreover, since all the values in a probability matrix are probabilities, they must be values in the interval $[0,1]$.

The $(k-1)$-dimensional \emph{probability simplex} (e.g., see~\cite{boyd2004convex}) is defined as
%\begin{equation}\label{def:simplex}
%\Delta^{(k-1)} \defeq \left\{\langle   x_1, \ldots, x_{k}\rangle \in \mathbb{R}^{k} \big| \sum_{ i=1}^{k} %x_{i} = 1 \land \bigwedge_{i = 1}^{k} x_{i}\geq 0\right\}.
%\end{equation}
\begin{equation}\label{def:simplex}
\Delta^{(k-1)} \defeq \left\{x \in \mathbb{R}_{\geq 0}^{k} \bigg| \sum_{ i=1}^{k} x_i = 1 \right\}
\end{equation}
where $x_i$ is the $i$-th component of vector $x$.
%\begin{equation}\label{def:simplex}
%\begin{aligned} \Delta^{(k-1)} \defeq &\left\{\langle   x_1, \ldots, x_{k}\rangle \in \mathbb{R}_{\geq 0}^{k} \bigg|\phantom{\sum_{ i=1}^{k}}\right.\\
%&\left.\phantom{\langle   x_1, \ldots, x_{k}\rangle \in}\sum_{ i=1}^{k} x_{i} = 1\right\}.\end{aligned}
%\end{equation}
The function $\gamma$ maps any point in $\Delta^{(n^2-1)}$ into a $n\times n$ probability matrix and every probability matrix is the image of 
a point of the same simplex, i.e., $\gamma(\Delta^{(n^2-1)})=\mathcal{P}_n$ and $\Delta^{(n^2-1)}=\gamma^{-1}(\mathcal{P}_n)$.
Hence, the set of the  
matrices $M_P \in \mathcal{P}_n$ such that $\sigma(M_P) < c$ corresponds 
to the set of points $x \in \Delta^{(n^2-1)}$ such that $\sigma(\gamma(x)) < c$, i.e., 
\begin{equation}
\mathcal{S}_{\sigma, n}(c) \defeq \left\{x \in \Delta^{(n^2-1)} \, \big|\, \sigma(\gamma(x)) < c \right\}.
\end{equation}

\begin{figure*}[!ht]
\begin{center}
     \begin{subfigure}[b]{0.49\textwidth}
 \resizebox{\textwidth}{!}{
\tikzset{every mark/.append style={scale=0.5}}
    \begin{tikzpicture}
    \begin{axis}
[   view={-45}{60},
    xmin=0,xmax=1,
    ymin=0,ymax=1,
    zmin=0, zmax=1,
    xlabel = {$x_1$},
    ylabel = {$x_2$},
    zlabel = {$x_3$}
    ]
	\addplot3[red, only marks] file{./out_set.txt};
	\addlegendentry{$\Delta^3 \setminus \mathcal{S}_{\IA,2}(0.3)$}
	\addplot3[blue, only marks] file{./in_set.txt};
	\addlegendentry{$\mathcal{S}_{\IA,2}(0.3)$}
    \end{axis}

    \end{tikzpicture}}
   \end{subfigure}
   \hfill
        \begin{subfigure}[b]{0.49\textwidth}
 \resizebox{\textwidth}{!}{
   \tikzset{every mark/.append style={scale=0.5}}
    \begin{tikzpicture}
    \begin{axis}
[   view={0}{0},
    xmin=0,xmax=1,
    ymin=0,ymax=1,
    zmin=0, zmax=1,
    xlabel = {$x_1$},
    ylabel = {$x_2$},
    zlabel = {$x_3$}
    ]
	\addplot3[red, only marks, mark=square*] file{./out_set.txt};
	\addlegendentry{$\Delta^3 \setminus \mathcal{S}_{\IA,2}(0.3)$}
	\addplot3[blue, only marks] file{./in_set.txt};
	\addlegendentry{$\mathcal{S}_{\IA,2}(0.3)$}
    \end{axis}

    \end{tikzpicture}}
   \end{subfigure}

\caption{A 3D graphical representation of the set $\mathcal{S}_{\IA, 2}(0.3)$. 
We uniformly sampled 2000 points in the cube $\Delta^{3}$ and plotted them. The blue-coloured 
points belong to $\mathcal{S}_{\IA, 2}(0.3)$, while
the red ones lays in $\Delta^{3} \setminus \mathcal{S}_{\IA, 2}(0.3)$.}\label{fig:S}
\end{center}
\end{figure*}
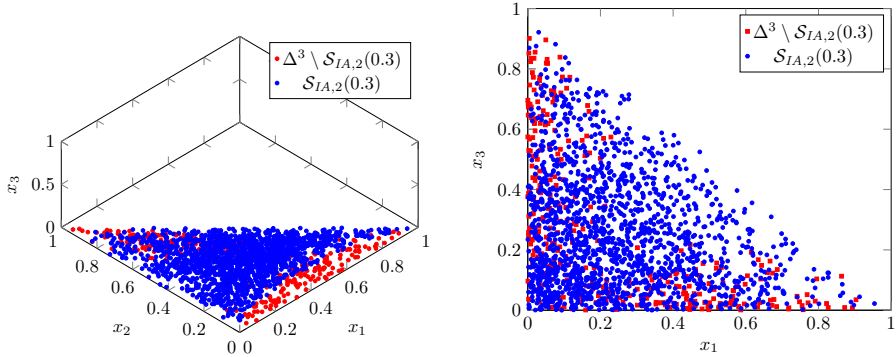

For example, Figure~\ref{fig:S} represents the set $\mathcal{S}_{\IA, 2}(0.3)$.

For any $\langle   x_1, \ldots, x_{n^2} \rangle \in \Delta^{(n^2-1)}$, $x_{n^2} = 1 - \sum_{ i=1}^{n^2-1} x_i$ according to Eq.~\ref{def:simplex}.
As a consequence, the function $\pi(\langle   x_1, \ldots, x_{n^2} \rangle) \defeq \langle   x_1, \ldots, x_{n^2-1} \rangle$ is bijective on $\Delta^{(n^2-1)}$ and any point in 
$\Delta^{(n^2-1)}$ corresponds to a point in $\pi(\Delta^{(n^2-1)}) \subset \mathbb{R}^{n^2-1}$.
%and there also exists a bijective function $\pi$ from $\Delta^{(n^2-1)}$ to a subset of $\mathbb{R}^{(n^2-1)}$, e.g., 
%$\pi(\langle   x_1, \ldots, x_{n^2} \rangle) \defeq \langle   x_1, \ldots, x_{n^2-1} \rangle$. Hence, $\rho_\sigma(c)$ is the ratio between the volumes of $\pi(\mathcal{S}_{\sigma}(c))$ and 
%$\pi(\Delta^{(n^2-1)})$ where $\pi(S) \defeq \{\pi(x)\,|\, x \in S\}$.
The Lebesgue measure of $\Delta^{(k-1)}$, i.e., its volume, on dimension $k-1$ is $\frac{1}{(k-1)!}$~\cite{simplex}. Hence, if $\mathcal{S}_{\sigma,n}(c)$ is Lebesgue-measurable, then 
\begin{equation}\label{def:rho}
\rho_{\sigma,n}(c) = \frac{V(\mathcal{S}_{\sigma,n}(c))}{V(\Delta^{(n^2-1)})}={(n^2-1)!}\, \int_{\Delta^{(n^2-1)}} \mathbf{1}_{\mathcal{S}_{\sigma,n}(c)}(x) d\, x
\end{equation}
where $V(S)$ denotes the Lebesgue measure of $S$ on dimension $(n^2-1)$. %$\pi(S) \defeq \{\pi(x)\,|\, x \in S\}$.
The value $\rho_{\sigma,n}(c)$ is the $\sigma$-significativity of $c$ in $\mathcal{P}_n$.

O-minimal sets are Lebesgue-measurable~\cite{Dries_1998}. Thus, when $\mathcal{S}_{\sigma,n}(c))$ is o-minimal, $\rho_{\sigma,n}(c)$ is well defined.
This is the case for $\sigma$ among Cohen's $\kappa$, Scott's $\pi$, Yule's $Y$, 
Fleiss's $\kappa$, $\IR$, and $\IA$. Even though their definitions include a division, $\mathcal{S}_{\sigma, n}(c)$, where $\sigma$ is any of the cited agreement measures, 
is definable in $(\mathbb{R},  \{0,1\}, +, $ $*, e^{x},  >)$ and is an o-minimal set.

\subsection{Evaluating $\rho_{\sigma, n}(c)$}

The analytic evaluation of the integral in Eq.~\ref{def:rho} is not always possible and relies on the form of $\sigma(\cdot)$.
In any case, we can numerically estimate it using the Monte Carlo integration method.
This method approximates the integral of a function $f: D \subseteq \mathbb{R}^n \rightarrow \mathbb{R}$ over $\Omega \subseteq D$ as 
\begin{equation}
\int_{\Omega} f(x) d\, x = \lim_{N \rightarrow +\infty} \frac{V(\Omega)}{N} \sum_{i=1}^{N} f(x_i)
\end{equation}
where $x_1, \ldots x_N$ are uniformly sampled point in $\Omega$  
(e.g., see~\cite{Mackay1998}). As in the discrete case, the approximation error is proportional to $1/\sqrt{N}$.

The following theorem suggests how to uniformly sample $\Delta^{(n^2-1)}$.
\begin{theorem}\cite[Ch.~5, Theorem~2.2]{devroye:1986}\label{theo:sampling_simplex}
If $E_1, \ldots, E_{k}$ are independent and identically distributed exponential random variables, then $\left\langle \frac{E_1}{\sum_{i=1}^{k}E_i}, \ldots, \frac{E_k}{\sum_{i=1}^{k}E_i} \right\rangle$ 
is uniformly distributed over $\Delta^{(k-1)}$.
\end{theorem}

Hence, we can approximate $V(\mathcal{S}_{\sigma}(c))$ as
\begin{equation}
V(\mathcal{S}_{\sigma,n}(c))= \int_{\Delta^{(n^2-1)}} \mathbf{1}_{\mathcal{S}_{\sigma,n}(c)}(x) d\, x \approx 
 \frac{V(\Delta^{(n^2-1)})}{N} \sum_{i=1}^{N} \mathbf{1}_{\mathcal{S}_{\sigma,n}(c)}(y_i),
\end{equation}
where $y_1, \ldots, y_N$ are uniformly distributed samples of $\Delta^{(n^2-1)}$.

From Eq.~\ref{def:rho}, it follows that:
\begin{equation}
\rho_{\sigma,n}(c) %= \frac{V(\mathcal{S}_{\sigma,n}(c))}{V(\Delta^{(n^2-1)})}
\approx \frac{1}{N} \sum_{i=1}^{N} \mathbf{1}_{\mathcal{S}_{\sigma,n}(c)}(y_i).
\end{equation}

\begin{figure*}[!ht]
     \begin{subfigure}[b]{0.49\textwidth}
\resizebox{\textwidth}{!}{
\begin{tikzpicture}
    \begin{axis}[xmin=0,xmax=1,
                        ymin=0,ymax=1,
                        xlabel={$c$},
                        ylabel={$\IA$–Significativity},
      			legend pos=south east,
			legend style={cells={align=left}}]
	\addplot table [x=IA,y=rho2,col sep=space] {IA_rho.csv};
	\addlegendentry{$\rho_{\IA,2}(c)$}
	\addplot table [x=IA,y=rho3,col sep=space] {IA_rho.csv};
	\addlegendentry{$\rho_{\IA,3}(c)$}
	\addplot table [x=IA,y=rho4,col sep=space] {IA_rho.csv};
	\addlegendentry{$\rho_{\IA,4}(c)$}
	\addplot table [x=IA,y=rho5,col sep=space] {IA_rho.csv};
	\addlegendentry{$\rho_{\IA,5}(c)$}
\end{axis}
\end{tikzpicture}}
\caption{The functions $\rho_{\IA,n}(c)$.}
\end{subfigure}
%\hfill
%     \begin{subfigure}[b]{0.31\textwidth}
%\resizebox{\textwidth}{!}{
%\begin{tikzpicture}
%    \begin{axis}[xmin=0,xmax=1,
%                        ymin=0,ymax=1,
%                        xlabel={$x$},
%                        ylabel={\% of $2\times 2$-matrices (Probability)},
%      			legend pos=south east,
%			legend style={cells={align=left}}]
%	\addplot table [x=ir,y=p,col sep=space] {IR_rho.csv};
%	\addlegendentry{$\rho_{\IR}(x)$}
%\end{axis}
%\end{tikzpicture}}
%\caption{The function $\rho_{\IR}(x)$.}
%\end{subfigure}
\hfill
\begin{subfigure}[b]{0.49\textwidth}
\resizebox{\textwidth}{!}{
\begin{tikzpicture}
    \begin{axis}[xmin=-1,xmax=1,
                        ymin=0,ymax=1,
                        xlabel={$c$},
                        ylabel={$\kappa$-Significativity},
      			legend pos=south east,
			legend style={cells={align=left}}]
	\addplot table [x=kappa,y=rho2,col sep=space] {kappa_rho.csv};
	\addlegendentry{$\rho_{\kappa,2}(c)$}
	\addplot table [x=kappa,y=rho3,col sep=space] {kappa_rho.csv};
	\addlegendentry{$\rho_{\kappa,3}(c)$}
	\addplot table [x=kappa,y=rho4,col sep=space] {kappa_rho.csv};
	\addlegendentry{$\rho_{\kappa,4}(c)$}
	\addplot table [x=kappa,y=rho5,col sep=space] {kappa_rho.csv};
	\addlegendentry{$\rho_{\kappa,5}(c)$}
\end{axis}
\end{tikzpicture}}
\caption{The functions $\rho_{\kappa,n}(c)$.}
\end{subfigure}
\caption{The $\IA$-significativity and Cohen's $\kappa$-significativity. When $\sigma$ is a agreement measure between classifiers having $n$ possible outcomes, 
the function 
$\rho_{\sigma,n}(x)$ is the ratio between the number of  $n\times n$-probability matrices $M \in \mathcal{P}_n$ such that $\sigma(M)< c$ and the 
overall number of $n\times n$-probability matrices.}\label{fig:rho}
\end{figure*}
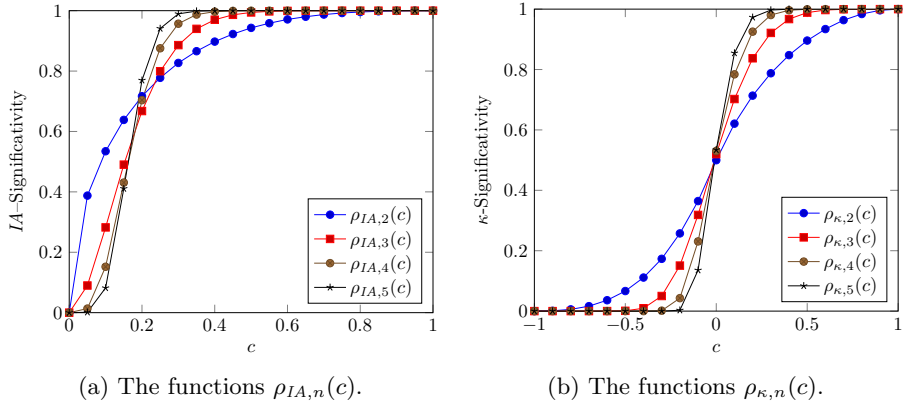

Sampling $\Delta^{(n^2-1)}$ according to Theorem~\ref{theo:sampling_simplex} takes time $\Theta(k)$ per 
sample. Thus, when $T(\sigma) \in \Theta(n^2)$, the 
Monte Carlo method can estimate $\rho_{\sigma,n}(c)$ in time $\Theta(n^2N)$.

\begin{example}
Let $M_C$ be the matrix as defined in Example~\ref{ex:varrho}. 
Cohen's $\kappa$ and $\IA$ for $M_P\defeq T_P(M_C)$ are $\kappa(M_P)=\kappa(M_C)\approx 0.70588$ and $\IA(M_P)=\IA(M_C)\approx 0.52115$, respectively.
The $\kappa$-significativity of $\kappa(M_P)$ in $\mathcal{P}_{2}$ is $\rho_{\kappa,2}(\kappa(M_P))\approx 0.9642$. 
Thus, more than $96\%$ of the $2\times 2$-probability matrices have a $\kappa$ value lower than that of $M_P$ and 
the probability of choosing by chance a probability matrix $M_P' \in \mathcal{P}_{2}$ with $\kappa(M_P') < \kappa(M_P)$ is 
$0.9642$.

Instead, the $\IA$-significativity of $\IA(M_P)$ in $\mathcal{P}_{2}$ is $\rho_{\IA,2}(\IA(M_P))\approx 0.9507$.
Hence, less than $5\%$ of the $2\times 2$-probability matrices have an $\IA$ value greater than or equal to that of $M_P$ and
the probability of choosing by chance a probability matrix $M_P' \in \mathcal{P}_{2}$ with $\IA(M_P') \leq \IA(M_P)$ is 
$0.9507$.
\end{example}

Figure~\ref{fig:rho} shows $\IA$ and Cohen's $\kappa$-significativity estimations as 
$n$ ranges between $2$ and $5$.

\subsection{Significativity Relation}

The following theorem relates the $\sigma$-significativities of $c$ in $\mathcal{P}_n$ and 
$\mathcal{M}_{n, m}$.

\begin{theorem}\label{theo:riemann}
If $\mathcal{S}_{\sigma,n}(c)$ is Riemann-measurable and $\sigma(M_C) = \sigma(T_P(M_C))$ for all $M_C \in \mathcal{M}_{n,m}$, then
\begin{equation}
\lim_{m\rightarrow +\infty}\varrho_{\sigma,n,m}(c) =\rho_{\sigma, n}(c).
\end{equation}
\end{theorem}
\begin{proof}
See Appendix~\ref{sec:proofs_riemann}.
\end{proof}

Since Riemann-measurability implies Lebesgue-measurability~\cite{Apostol:105425}, $\rho_{\sigma, n}(c)$ is well defined 
when $\mathcal{S}_{\sigma,n}(c)$ is Riemann-measurable.
It follows that, when $\mathcal{S}_{\sigma,n}(c)$ is Riemann-measurable and 
$\sigma(M_C) = \sigma(T_P(M_C))$, 
$\rho_{\sigma, n}(c)$ can be used as an approximation of $\varrho_{\sigma,n,m}(c)$.

Although the Riemann-integrability may seem a restrictive condition, the following results prove that it is quite commonly satisfied, and every bounded o-minimal set is integrable by
Riemann.

\begin{lemma}\label{lemma:ominimal}
Every bounded o-minimal subset of $\mathbb{R}^n$ is Riemann-measurable.
\end{lemma}
\begin{proof}
A bounded set $S$ is Riemann-measurable if and only if its indicator function $\mathbf{1}_{S}$ is Riemann integrable. 
By Lebesgue's criterion for Riemann integrability (e.g., see~\cite[Theorem~14.5]{Apostol:105425}), any function $f$ definable and bounded on a compact interval 
$I\subseteq \mathbb{R}^n$ is 
Riemann-integrable on $I$ if and only if the set of discontinuities of $f$ in $I$ has $n$-dimensional Lebesgue measure $0$.
However, the set of discontinuities of  $\mathbf{1}_{S}$ matches the topological boundary of $S$.
If $S \subseteq \mathbb{R}^n$ is o-minimal, then the dimension of the boundary of $S$ is lower than $m$ in $\mathbb{R}^n$~\cite[Ch.~4, Corollary 1.10]{Dries_1998}.
Thus, the boundary of $S$ has $m$-dimensional Lebesgue measure $0$, and $S$ is Riemann-measurable.
\end{proof}

The following corollary is a direct consequence of Theorem~\ref{theo:riemann} and Lemma~\ref{lemma:ominimal}.

\begin{corollary}\label{cor:ominimal}
If $\mathcal{S}_{\sigma, n}(c)$ is definable in an o-minimal theory over $\mathbb{R}$ and $\sigma(M_C) = \sigma(T_P(M_C))$ for all $M_C \in \mathcal{M}_{n,m}$, then
\begin{equation}
\lim_{m\rightarrow +\infty} \varrho_{\sigma,n,m}(c) =\rho_{\sigma, n}(c).
\end{equation}
\end{corollary}

It is easy to verify that $\sigma(M_C) = \sigma(T_P(M_C))$ holds for Cohen's $\kappa$ and $\IA$. Since, as pointed out in this section,
$\mathcal{S}_{\sigma,n}(c)$ is an o-minimal set for the same agreement measures, Corollary~\ref{cor:ominimal}'s thesis holds for them.

Figure~\ref{fig:error} shows the estimated errors in approximating $\varrho_{\sigma,n,m}(c)$ by $\rho_{\sigma, n}(c)$ as $m$ changes for $\sigma$ among Cohen's $\kappa$ and $\IA$.
As expected, the difference between the two indices decreases as $m$ increases. 
However, when the data set consists of 10 entries, it exceeds 0.8 and 0.3 for $\IA$ and Cohen's $\kappa$, respectively. Thus, if the data set is not large enough, then $\rho_{\sigma, n}(c)$ can not effectively approximate $\varrho_{\sigma,n,m}(c)$.

%
%The semi-algebraic theory of the polynomials on $\mathbb{R}$, i.e., $(\mathbb{R},  \{0,1\}, +, $ $*, >)$ and  
%its Pfaffian extensions~\cite{speissegger1999pfaffian}, e.g.,  $(\mathbb{R},  \{0,1\}, +, $ $*,e^{x},  >)$ or $(\mathbb{R},  \{0,1\}, +, $ $*,\ln{x},  >)$, are o-minimal theories. 
%Thus, the above results hold for Cohen's $\kappa$, $\IR$, $\IA$, and all statistic coefficients definable as rational functions on $\mathbb{R}$.

%If $\rho_{f}(c)$ is computable, then we can define a quality scale for $\sigma(\cdot)$ by sampling over all the admissible values of $c$, i.e., $[0,1]$, or by using a dichotomic search~\cite{Cormen:2009}. For instance, we may identify the value $c_0, c_1, c_2, c_3$ such that 
%$\rho_{f}(c_0)=0.2$, $\rho_{f}(c_1)=0.4$, $\rho_{f}(c_2)=0.6$, and $\rho_{f}(c_3)=0.8$, and, miming the McHugh's scale for Cohen's $\kappa$~\cite{McHugh:2012}, 
%use them as boundaries of the scale classes, e.g., 
%$[0, c_0)$ (``none to slight''), $[c_0, c_1)$ (``fair''), $[c_1, c_2)$ (``moderate''),
%$[c_2, c_3)$ (``substantial''), and $[c_3, 1.0)$ (``perfect or almost perfect agreement'').

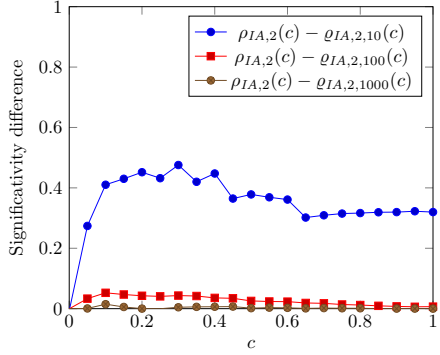
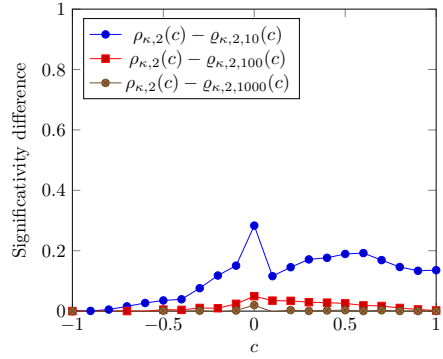
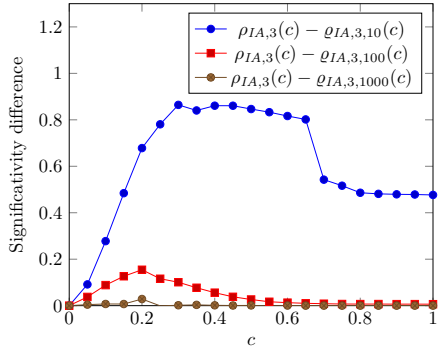
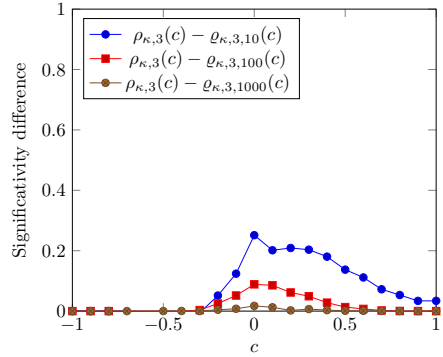
\begin{figure*}[!ht]
\begin{subfigure}[b]{0.49\textwidth}
\resizebox{\textwidth}{!}{
\begin{tikzpicture}
    \begin{axis}[ xmin=0,xmax=1,
                        ymin=-0.1,ymax=1,
                        xlabel={$c$},
                        ylabel={Significativity difference},
      			legend pos=north east,
			legend style={cells={align=left}}]
	%\addplot table [x=v,y=delta2_2,col sep=space] {IA_delta.csv};
	%\addlegendentry{$\rho_{\IA,2}(x)-\varrho_{\IA, 2, 2}(x)$}
	%\addplot table [x=IA,y=delta2_4,col sep=space] {IA_delta.csv};
	%\addlegendentry{$\rho_{\IA,2}(x)-\varrho_{\IA, 2, 4}(x)$}
	%\addplot table [x=IA,y=delta2_8,col sep=space] {IA_delta.csv};
	%\addlegendentry{$\rho_{\IA,2}(x)-\varrho_{\IA, 2, 8}(x)$}
	\addplot table [x=IA,y=delta2_10,col sep=space] {IA_delta.csv};
	\addlegendentry{$\rho_{\IA,2}(c)-\varrho_{\IA, 2, 10}(c)$}
	\addplot table [x=IA,y=delta2_100,col sep=space] {IA_delta.csv};
	\addlegendentry{$\rho_{\IA,2}(c)-\varrho_{\IA, 2, 100}(c)$}
	\addplot table [x=IA,y=delta2_1000,col sep=space] {IA_delta.csv};
	\addlegendentry{$\rho_{\IA,2}(c)-\varrho_{\IA, 2, 1000}(c)$}
\end{axis}
\end{tikzpicture}}
\caption{The difference between $\rho_{\IA, 2}(c)$ and 
$\varrho_{\IA, 2, m}(c)$.}
\end{subfigure}
%\hfill
%     \begin{subfigure}[b]{0.31\textwidth}
%\resizebox{\textwidth}{!}{
%\begin{tikzpicture}
%    \begin{axis}[xmin=0,xmax=1,
%                        ymin=0,ymax=1,
%                        xlabel={$x$},
%                        ylabel={\% of $2\times 2$-matrices (Probability)},
%      			legend pos=south east,
%			legend style={cells={align=left}}]
%	\addplot table [x=ir,y=p,col sep=space] {IR_rho.csv};
%	\addlegendentry{$\rho_{\IR}(x)$}
%\end{axis}
%\end{tikzpicture}}
%\caption{The function $\rho_{\IR}(x)$.}
%\end{subfigure}
\hfill
\begin{subfigure}[b]{0.49\textwidth}
\resizebox{\textwidth}{!}{
\begin{tikzpicture}
    \begin{axis}[ xmin=-1,xmax=1,
                        ymin=-0.1,ymax=0.3,
                        xlabel={$c$},
                        ylabel={Significativity difference},
      			legend pos=north west,
			legend style={cells={align=left}}]
	\addplot table [x=kappa,y=delta2_10,col sep=space] {kappa_delta.csv};
	\addlegendentry{$\rho_{\kappa,2}(c)-\varrho_{\kappa, 2, 10}(c)$}
	\addplot table [x=kappa,y=delta2_100,col sep=space] {kappa_delta.csv};
	\addlegendentry{$\rho_{\kappa,2}(c)-\varrho_{\kappa, 2, 100}(c)$}
	\addplot table [x=kappa,y=delta2_1000,col sep=space] {kappa_delta.csv};
	\addlegendentry{$\rho_{\kappa,2}(c)-\varrho_{\kappa, 2, 1000}(c)$}
\end{axis}
\end{tikzpicture}}
\caption{The difference between $\rho_{\kappa, 2}(c)$ and 
$\varrho_{\kappa, 2, m}(c)$.}
\end{subfigure}
\begin{subfigure}[b]{0.49\textwidth}
\resizebox{\textwidth}{!}{
\begin{tikzpicture}
    \begin{axis}[ xmin=0,xmax=1,
                        ymin=-0.1,ymax=1.0,
                        xlabel={$c$},
                        ylabel={Significativity difference},
      			legend pos=north east,
			legend style={cells={align=left}}]
	%\addplot table [x=v,y=delta3_2,col sep=space] {IA_delta.csv};
	%\addlegendentry{$\rho_{\IA,3}(x)-\varrho_{\IA, 3, 2}(x)$}
	%\addplot table [x=IA,y=delta3_4,col sep=space] {IA_delta.csv};
	%\addlegendentry{$\rho_{\IA,3}(x)-\varrho_{\IA, 3, 4}(x)$}
	%\addplot table [x=IA,y=delta3_8,col sep=space] {IA_delta.csv};
	%\addlegendentry{$\rho_{\IA,3}(x)-\varrho_{\IA, 3, 8}(x)$}
	\addplot table [x=IA,y=delta3_10,col sep=space] {IA_delta.csv};
	\addlegendentry{$\rho_{\IA,3}(c)-\varrho_{\IA, 3, 10}(c)$}
	\addplot table [x=IA,y=delta3_100,col sep=space] {IA_delta.csv};
	\addlegendentry{$\rho_{\IA,3}(c)-\varrho_{\IA, 3, 100}(c)$}
	\addplot table [x=IA,y=delta3_1000,col sep=space] {IA_delta.csv};
	\addlegendentry{$\rho_{\IA,3}(c)-\varrho_{\IA, 3, 1000}(c)$}
\end{axis}
\end{tikzpicture}}
\caption{The difference between $\rho_{\IA, 3}(c)$ and 
$\varrho_{\IA, 3, m}(c)$.}
\end{subfigure}
%\hfill
%     \begin{subfigure}[b]{0.31\textwidth}
%\resizebox{\textwidth}{!}{
%\begin{tikzpicture}
%    \begin{axis}[xmin=0,xmax=1,
%                        ymin=0,ymax=1,
%                        xlabel={$x$},
%                        ylabel={\% of $2\times 2$-matrices (Probability)},
%      			legend pos=south east,
%			legend style={cells={align=left}}]
%	\addplot table [x=ir,y=p,col sep=space] {IR_rho.csv};
%	\addlegendentry{$\rho_{\IR}(x)$}
%\end{axis}
%\end{tikzpicture}}
%\caption{The function $\rho_{\IR}(x)$.}
%\end{subfigure}
\hfill
\begin{subfigure}[b]{0.49\textwidth}
\resizebox{\textwidth}{!}{
\begin{tikzpicture}
    \begin{axis}[ xmin=-1,xmax=1,
                        ymin=-0.1,ymax=0.3,
                        xlabel={$c$},
                        ylabel={Significativity difference},
      			legend pos=north west,
			legend style={cells={align=left}}]
	\addplot table [x=kappa,y=delta3_10,col sep=space] {kappa_delta.csv};
	\addlegendentry{$\rho_{\kappa,3}(c)-\varrho_{\kappa, 3, 10}(c)$}
	\addplot table [x=kappa,y=delta3_100,col sep=space] {kappa_delta.csv};
	\addlegendentry{$\rho_{\kappa,3}(c)-\varrho_{\kappa, 3, 100}(c)$}
	\addplot table [x=kappa,y=delta3_1000,col sep=space] {kappa_delta.csv};
	\addlegendentry{$\rho_{\kappa,3}(c)-\varrho_{\kappa, 3, 1000}(c)$}
\end{axis}
\end{tikzpicture}}
\caption{The difference between $\rho_{\kappa, 3}(c)$ and 
$\varrho_{\kappa, 3, m}(c)$.}
\end{subfigure}
\caption{The difference between $\rho_{\sigma, n}(c)$ and 
$\varrho_{\sigma, n, m}(c)$ for $m \in \{10, 100, 1000\}$ when $\sigma$ is 
 Cohen's $\kappa$ and $\IA$.}\label{fig:error}
\end{figure*}

\section{Conclusion}\label{sec:conclusions}

%Rating an agreement measure can be challenging, especially when dealing with various effect sizes. 

%Moreover, we have shown that the absence of any absolute qualitative reference scale for the agreement measures plays a role in the interpretation of their statistical significance.

This work introduces a general technique to evaluate the statistical relevance of agreement values.
Our proposal does not gauge the meaningfulness of the data set used to build a confusion matrix. It instead evaluates the significativity of an agreement value over a data set once the data set size has been set.
We introduced the $\sigma$-significativity of $c$ 
over $n\times n$ confusion matrices that collect $m$ 
classifications, $\varrho_{\sigma,n,m}(c)$, as the probability of choosing by chance a 
confusion matrix having an agreement value lower than $c$.
This measure is parametetric in the agreement measure $\sigma$, the number of classes $n$, and
the size of the data set $m$.
We also define the $\sigma$-significativity of $c$ 
over $n\times n$ probability matrices, $\rho_{\sigma,n}(c)$, as the probability of 
choosing by chance a probability matrix having an agreement value lower than $c$.
As long as the set of the probability matrices whose agreement value is lower than $c$, $\mathcal{S}_{\sigma, n}(c)$, is definable in an o-minimal theory, the two $\sigma$-significativity 
converge as $m$ tends to infinity.
%The computability of the two significativity measures is also addressed.

The $\sigma$-significativity over confusion matrices is computable. However, the asymptotic time complexity of its exact evaluation is so high that it discourages its use. Hence, we suggested a Monte Carlo numerical estimator for it whose complexity is linear in $m$ and quadratic in $n$.
On the other hand, the $\sigma$-significativity over probability matrices depends on $\sigma$ and, in some cases, it is not analytically computable. We proposed a numerical method, with quadratic time complexity in $n$, to estimate this index too. The algorithms have been implemented in \texttt{R} package, named \texttt{rSignificativity}, which is available on \href{https://albertocasagrande.github.io/rSignificativity}{GitHub}. This package was used in combination with PGF/TikZ~\cite{tantau2018pgf} to produce the figures in this manuscript. 
%The package is also available to the public as an online tool on \href{https://albertocasagrande.github.io/significativity/}. The online tool will make it easier for researchers who do not have a programming background to produce publishable results.

The notion of $\sigma$-significativity is meant to provide a 
statistical significance to the agreement values and not to replace 
them. In this spirit, we plan to investigate the relation
between agreement scales, such as the one proposed by Landis and Kock~\cite{Landies1977}, and the $\sigma$-significativity.

We also plan to use $\sigma$-significativity in both AI and clinical domain.
For example, in training AI algorithms for medical image classifications, a parametric scale for any agreement value could assess the consistency between human annotators and algorithmic predictions.

\bibliographystyle{plain}
\bibliography{IT_biblio}

\appendix

\section{Proof of Theorem~\ref{theo:riemann}}\label{sec:proofs_riemann}

While the function $T_P$ maps any confusion matrix into a probability matrix, some probability matrices do not correspond to any confusion matrix because $T_P$ is not bijective. 
It is easy to see that, for any confusion matrix $M_C \in \mathbb{N}^{n \times n}$, $T_P(M_C) \in \mathbb{Q}^{n \times n}$. Thus, 
the probability matrices with some irrational components are not images of any confusion matrix.

Let $\mathcal{P}_{n, m}$ be the set of probability matrices corresponding to a $n\times n$ confusion matrix whose components 
sum up to $m$, i.e.,
\begin{equation}\label{eq:prob_n_m}
\mathcal{P}_{n, m} \defeq T_P\left(\mathcal{M}_{n,m} \right).
\end{equation}
Since $\gamma: \Delta^{(n^2-1)} \rightarrow \mathcal{P}_{n}$ is bijective, we can also define the set
\begin{equation}\label{eq:pC_n_m}
\mathcal{C}_{m,n^2}^*\defeq \gamma^{-1}\left(\mathcal{P}_{n, m} \right).
\end{equation}
By construction, $\mathcal{P}_{n, m}$ is a set of $n\times n$ rational probability matrices (see Eq.~\ref{def:T_P}), 
i.e., $\mathcal{P}_{n, m} \subset \mathcal{P}_{n} \cap \mathbb{Q}^{n \times n}$. Hence, 
$\mathcal{C}_{m,n^2}^* = \gamma^{-1}\left(\mathcal{P}_{n, m} \right) \subset  \gamma^{-1}\left(\mathcal{P}_{n} \right) = \Delta^{(n^2-1)}$
and $\pi(\mathcal{C}_{m,n^2}^*) = \pi(\Delta^{(n^2-1)})$. It is worth to notice that both $\mathcal{P}_{n,m}$ and $\mathcal{C}_{m,n^2}^*$ 
have finite cardinalities because
$T_P: \mathcal{M}_{n,m} \rightarrow \mathcal{P}_{n,m}$ and $\gamma: \mathcal{P}_{n} \rightarrow \mathcal{C}_{m,n^2}$
are bijective. In particular, $|\mathcal{C}_{m,n^2}^*| = |\mathcal{P}_{n,m}| = |\mathcal{M}_{n,m}| = |\mathcal{C}_{m,n^2}|=\binom{m+n^2-1}{m}$.

The matrices in $\mathcal{P}_{n, m}$ correspond to evenly spread points in $\pi\left( \Delta^{(n^2-1)} \right)$. 
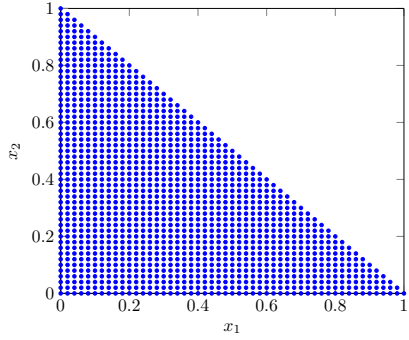
\begin{figure}[!ht]
\begin{center}
\tikzset{every mark/.append style={scale=0.5}}
\resizebox{0.45\textwidth}{!}{%
    \begin{tikzpicture}
    \begin{axis}
[   view={0}{90},
    xmin=0,xmax=1,
    ymin=0,ymax=1,
    zmin=0, zmax=1,
    xlabel = {$x_1$},
    ylabel = {$x_2$},
    zlabel = {$x_3$}
    ]
	\addplot3[blue, only marks] file{./confusion_matrix_2x2_proj.txt};
    \end{axis}

    \end{tikzpicture}}
    \caption{A $2$-dimensional projection of the points in $\pi(\mathcal{C}_{2, 50}^*)$ (the blue set). The points are evenly spread inside $\pi(\Delta^{3})$.
    The minimal distance between two distinct points $x, x' \in \pi(\mathcal{C}_{2, 50}^*)$ is $1/50$.}\label{fig:confusion_matrices}
\end{center}
\end{figure}
These points 
induce a uniform grid with cell side length $1/m$ such that each of the cells included in $\pi\left( \Delta^{(n^2-1)} \right)$ contains one of the points. 

\begin{lemma}\label{lemma:distance}
The smallest distance between two distinct vectors in $\pi\left(\mathcal{C}_{m,n^2}^*  \right)$ is $1/m$. %, i.e., 
%\begin{equation*}
%\min_{y,y' \in \pi\left(\gamma^{-1}\left(\mathcal{P}_{n, m} \right)  \right)\setminus \text{Id}}\Mdist{y}{y'} = \frac{1}{m}.
%\end{equation*}
%
Moreover, if $y,y' \in \pi\left(\mathcal{C}_{m,n^2}^* \right)$ and $\|y-y'\|=1/m$, then all the components 
of $y-y'$, but one equal $0$. %, i.e., if $\langle x_1, \ldots, x_{n^2-1}\rangle =y -y'$, then there exists $i \in [1, n^2-1]$ such that 
%$|x_i| = 1/m$ and $x_j=0$ for all $j \neq i \in [1, n^2-1]$.
\end{lemma}
\begin{proof}
By construction, any point $\gamma\left(\mathcal{C}_{n,m}\right)$ has the form $\left\langle \frac{c_1}{m}, \ldots \frac{c_{n^2-1}}{m} \right\rangle$ with $c_i \in \mathbb{N}$ and $\sum_{i=1}^{n^2-1}c_i\leq m$. It follows that $c_i \in [1, m]$ for any $i\in [1, n^2-1]$.

If $y=\left\langle \frac{c_1}{m}, \ldots \frac{c_{n^2-1}}{m} \right\rangle$ and $y'=\left\langle \frac{c_1'}{m}, \ldots \frac{c_{n^2-1}'}{m} \right\rangle$, then 
\begin{align*}
\Mdist{y}{y'} &= \sqrt{\sum_{j=1}^{n^2-1} \left(\frac{c_j}{m}-\frac{c_j'}{m}\right)^2}= \sqrt{\frac{1}{m^2}\sum_{j=1}^{n^2-1} \left(c_j-c_j'\right)^2}
=\frac{1}{m}\sqrt{\sum_{j=1}^{n^2-1} \left(c_j-c_j'\right)^2}
\end{align*}
Hence, $y \neq y'$ if and only if $c_j \neq c_j'$ for some $j \in [1, n^2-1]$. 
Since both $c_j$ and $c_j'$ are natural numbers, if $c_j \neq c_j'$, then $ \left|c_j-c_j'\right| \geq 1$ and $\left(c_j-c_j'\right)^2 \geq 1$.
Thus, 
\begin{align*}
\Mdist{y}{y'} &=\frac{1}{m}\sqrt{\sum_{j=1}^{n^2-1} \left(c_j-c_j'\right)^2} \geq \frac{1}{m}\sqrt{k},
\end{align*}
where $k$ is the number of indices $j \in [1, n^2-1]$ such that $c_j \neq c_j'$, i.e., $k \defeq |D|$ where $D \defeq \{j \in [1, n^2-1]\, |\, c_j \neq c_j' \}$.
It follows that the minimum of $\Mdist{y}{y'}$ for $y\neq y'$ equals $1/m$. Moreover, 
if  $\Mdist{y}{y'}=1/m$, then $k=1$, i.e., $1=|D|$. Hence, $|[1, n^2-1] \setminus D| = n^2-2$ and $c_j = c_j'$ for all $j \in [1, n^2-1] \setminus D$. 
As a consequence,  $c_j - c_j' = 0$ for $n^2-2$ different indices $j \in [1, n^2-1] \setminus D$ and $|c_j - c_j'| = 1$ for the only $j \in D$.
\end{proof}

Let $\mathcal{R}_{\sigma,n,m}^*(c)$ be the subset of the probability matrices in $\mathcal{S}_{\sigma,n}(c)$ that correspond to a
$n\times n$ confusion matrix with $m$ tests, i.e., 
\begin{equation}\label{eq:pR_n_m}
\mathcal{R}_{\sigma,n,m}^*(c) \defeq \mathcal{S}_{\sigma,n}(c) \cap \mathcal{C}_{m,n^2}^*.
\end{equation}

\begin{lemma}\label{lemma:sigma}
Let $\sigma$ such that $\sigma(M_C) = \sigma(T_P(M_C))$ for all $M_C \in \mathcal{M}_{n,m}$. The sets $\mathcal{R}_{\sigma,n,m}^*(c)$ and  $\mathcal{R}_{\sigma,n,m}(c)$ have the same cardinality.
\end{lemma}
\begin{proof}
Since $\mathcal{R}_{\sigma,n,m}^*(c) \defeq \mathcal{S}_{\sigma,n}(c) \cap \mathcal{C}_{m,n^2}^*$, 
\begin{align*}
\left|\mathcal{R}_{\sigma,n,m}^*(c)\right| &=  \left|\mathcal{S}_{\sigma,n}(c) \cap \mathcal{C}_{m,n^2}^* \right|\\
&=  \left|\left\{x \in \Delta^{(n^2-1)} \,|\, \sigma(\gamma(x)) < c \right\} \cap \mathcal{C}_{m,n^2}^* \right|\\
&=  \left|\left\{x \in \mathcal{C}_{m,n^2}^* \,|\, \sigma(\gamma(x)) < c \right\}\right|.
\end{align*}
Moreover
\begin{align*}
\left|\mathcal{R}_{\sigma,n,m}^*(c)\right| &= \left|\left\{x \in \mathcal{C}_{m,n^2}^* \,|\, \sigma(\gamma(x)) < c \right\}\right|\\
&=  \left|\left\{x \in \gamma^{-1}\left(\mathcal{P}_{n,m}\right) \,|\, \sigma(\gamma(x)) < c \right\}\right|
\end{align*}
because $\mathcal{C}_{m,n^2}^* \defeq \gamma^{-1}\left(\mathcal{P}_{n,m}\right)$.
Hence, 
\begin{align*}
\left|\mathcal{R}_{\sigma,n,m}^*(c)\right| &=  \left|\left\{x \in \gamma^{-1}\left(\mathcal{P}_{n,m}\right) \,|\, \sigma(\gamma(x)) < c \right\}\right|\\
&=  \left|\left\{M_P \in \mathcal{P}_{n,m}\,|\, \sigma(M_P) < c \right\}\right|\\
&=  \left|\left\{M_P \in T_P\left(\mathcal{M}_{n,m}\right)\,|\, \sigma(M_P) < c \right\}\right|
\end{align*}
because $\gamma: P_n \rightarrow \Delta^{(n^2-1)}$ is bijective and $\mathcal{P}_{n,m} \defeq T_P\left(\mathcal{M}_{n,m}\right)$.

Since $T_P: \mathcal{M}_{n,m} \rightarrow \mathcal{P}_{n,m}$ is bijective,
\begin{align*}
\left|\mathcal{R}_{\sigma,n,m}^*(c)\right|
&=  \left|\left\{M_P \in T_P\left(\mathcal{M}_{n,m}\right)\,|\, \sigma(M_P) < c \right\}\right|\\
&=  \left|\left\{M_C \in \mathcal{M}_{n,m}\,|\, \sigma(T_P(M_C)) < c \right\}\right|.
\end{align*}

However,
\begin{align*}
\left|\mathcal{R}_{\sigma,n,m}^*(c)\right|
&=  \left|\left\{M_C \in \mathcal{M}_{n,m}\,|\, \sigma(T_P(M_C)) < c \right\}\right|\\
&=  \left|\left\{M_C \in \mathcal{M}_{n,m}\,|\, \sigma(M) < c \right\}\right|
\end{align*}
because $\sigma(T_P(M))=\sigma(M)$ for any $M_C \in \mathcal{M}_{n,m}$.

Finally,
\begin{align*}
\left|\mathcal{R}_{\sigma,n,m}^*(c)\right|
&=  \left|\left\{M_C \in \mathcal{M}_{n,m}\,|\, \sigma(M) < c \right\}\right|\\
&=  \left|\left\{M_C \in \gamma^{-1}\left(\mathcal{C}_{n,m}\right) \,|\, \sigma(M_C) < c \right\}\right|\\
&=  \left|\left\{x \in \mathcal{C}_{n,m} \,|\, \sigma(\gamma(x)) < c \right\}\right|\\
&= \left|\mathcal{R}_{\sigma,n,m}(c)\right|.
\end{align*}
Thus, the thesis holds.
\end{proof}

%Since both $\mathcal{P}_{n,m}$ is a finite set and $|\mathcal{P}_{n,m}|=|\mathcal{C}_{m,n^2}^*|$, we can define the ratio between 
%the number of probability matrices $M \in \mathcal{P}_{n,m}$ such that $\sigma(\gamma(M)) < c$ and $|\mathcal{P}_{n,m}|$ as:
%\begin{equation}
%\varrho_{\sigma,n,m}(c)^*=\frac{\left|\mathcal{R}_{\sigma,n,m}^*(c)\right|}{\left|\mathcal{C}_{m,n^2}^*\right|}
%\end{equation}

\begin{lemma}\label{lemma:riemann}
If $A \subseteq \Delta^{(n^2-1)}$ is Riemann-measurable in dimension $n^2-1$, then
\begin{equation}
V(A) = \lim_{m \rightarrow +\infty} \frac{1}{m^{n^2-1}} \left|A \cap \pi\left(\mathcal{C}_{m,n^2}^*\right)\right|
\end{equation}
\end{lemma}
\begin{proof}
Let us consider the $(n^2-1)$-dimensional grid of $[0,1]^{n^2-1}$ having the cells
\begin{equation}\label{eq:grid}
Q_{c_1, \ldots, c_{n^2-1}} \defeq  \prod_{i=1}^{n^2-1} \left[ \frac{2 c_i-1}{2m}, \frac{2 c_i + 1}{2m} \right]
\end{equation}
where $c_i \in [0, m]$ for any $i \in [1, n^2-1]$. Every cell $Q_{c_1, \ldots, c_{n^2-1}}$ has Lebesgue measure $1/m^{n^2-1}$ and 
its interior, $\interior{Q_{c_1, \ldots, c_{n^2-1}}}$, contains
$\langle c_1/m, \ldots, c_{n^2-1}/m\rangle$.

Since $T_P\left(\mathcal{M}_{n,m}\right)=\mathcal{P}_{n,m}$, %because of Eq.~\ref{eq:prob_n_m} 
$\gamma\left(\mathcal{C}_{m,n^2}\right)=\mathcal{M}_{n,m}$, and $\mathcal{C}_{m,n^2}^*=\gamma^{-1}\left(\mathcal{P}_{n,m}\right)$, 
the vector $\langle c_1, \ldots, c_{n^2-1}\rangle$ belongs to $\pi\left(\mathcal{C}_{m,n^2}\right)$ if and only if  
$\pi\left(\mathcal{C}_{m,n^2}^*\right)$ contains $\langle c_1/m, \ldots, c_{n^2-1}/m\rangle$.
Moreover,  the maximal distance between the vector $\langle c_1/m, \ldots, c_{n^2-1}/m\rangle$ and any point on the border of 
$Q_{c_1, \ldots, c_{n^2-1}}$ is $1/(\sqrt{2} m)$, and Lemma~\ref{lemma:distance} proves that the minimal distance 
between two distinct vectors in $\pi\left(\mathcal{C}_{m,n^2}^*\right)$ is $1/m$.
Hence, $\langle c_1, \ldots, c_{n^2-1}\rangle \in \pi\left(\mathcal{C}_{m,n^2}\right)$ if and only if the set
$Q_{c_1, \ldots, c_{n^2-1}} \cap \pi\left(\mathcal{C}_{m,n^2}^*\right)$ is the singleton $\{\langle c_1/m, \ldots, c_{n^2-1}/m\rangle\}$.
Moreover, $\langle c_1, \ldots, c_{n^2-1}\rangle \not\in \pi\left(\mathcal{C}_{m,n^2}\right)$ if and only if 
$Q_{c_1, \ldots, c_{n^2-1}} \cap \pi\left(\mathcal{C}_{m,n^2}^*\right)=\emptyset$.

For any set Riemann-measurable set $A \subseteq \pi(\Delta^{(n^2-1)})$,  we can define the union, $\overline{J}(A,m)$, 
of the cells $Q_{c_1, \ldots, c_{n^2-1}}$ that are not disjoint from $A$, i.e,
$$
\overline{J}(A,m) \defeq \bigcup_{ A \cap Q_{c_1, \ldots, c_{n^2-1}} \neq 0}  Q_{c_1, \ldots, c_{n^2-1}}.
$$
The interior of $\overline{J}(A,m)$ over-approximates $A$, i.e., $\interior{\overline{J}(A,m)} \supseteq A$. 

Since every cell $Q_{c_1, \ldots, c_{n^2-1}}$ has Lebesgue measure $1/m^{n^2-1}$ and contains exactly one 
of the vectors in $\pi\left(\mathcal{C}_{m,n^2}^*\right)$, the Lebesgue measure of $\overline{J}(A,m)$ 
is $1/m^{n^2-1}$ multiplied by the number of vectors in $\pi\left(\mathcal{C}_{m,n^2}^*\right) \cap \overline{J}(A,m)$, i.e.,
\begin{align*}
V(\overline{J}(A,m)) &= \frac{1}{m^{n^2-1}}\sum_{ A \cap Q_{c_1, \ldots, c_{n^2-1}} \neq 0}  1= \frac{1}{m^{n^2-1}} \left|\overline{J}(A,m) \cap \pi\left(\mathcal{C}_{m,n^2}^*\right)\right|.
\end{align*}
Since $\overline{J}(A,m) \supseteq A$, $A \cap \pi\left(\mathcal{C}_{m,n^2}^*\right)$ is a subset of $\overline{J}(A,m) \cap \pi\left(\mathcal{C}_{m,n^2}^*\right)$ and 
$V(\overline{J}(A,m)) \geq 1/m^{n^2-1} \left|A \cap \pi\left(\mathcal{C}_{m,n^2}^*\right)\right|$.

Analogously, 
the union, $\underline{J}(A,m)$, of the cells $Q_{c_1, \ldots, c_{n^2-1}}$ that are subsets of the interior of $A$, i.e,
$$
\underline{J}(A,m) \defeq \bigcup_{ \interior{A} \subseteq Q_{c_1, \ldots, c_{n^2-1}}}  Q_{c_1, \ldots, c_{n^2-1}},
$$
then $\underline{J}(A,m)$ under-approximates $A$, i.e.,  $\underline{J}(A,m) \subseteq A$. 

Since $\interior{Q_{c_1, \ldots, c_{n^2-1}}}$ contains
$\langle c_1/m, \ldots, c_{n^2-1}/m\rangle$, 
the Lebesgue measure of $\underline{J}(A,m)$ 
is $1/m^{n^2-1}$ multiplied by the number of vectors in $\pi\left(\mathcal{C}_{m,n^2}^*\right)$ 
that also belong to $\underline{J}(A,m)$, i.e.,
\begin{align*}
V(\underline{J}(A,m)) &= \frac{1}{m^{n^2-1}}\sum_{ \interior{A} \subseteq  Q_{c_1, \ldots, c_{n^2-1}}}  1 = \frac{1}{m^{n^2-1}} \left|\underline{J}(A,m) \cap \pi\left(\mathcal{C}_{m,n^2}^*\right)\right|.
\end{align*}
Since $\underline{J}(A,m) \subseteq A$, $\underline{J}(A,m) \cap \pi\left(\mathcal{C}_{m,n^2}^*\right)$ is a subset of  $A \cap \pi\left(\mathcal{C}_{m,n^2}^*\right)$ and 
$V(\underline{J}(A,m)) \leq 1/m^{n^2-1} \left|A \cap \pi\left(\mathcal{C}_{m,n^2}^*\right)\right|$.

We know that $\underline{J}(A,m) \subseteq A \subseteq \interior{\overline{J}(A,m)}$. Thus, $V(\underline{J}(A,m)) \leq V(A) \leq V(\overline{J}(A,m))$. As a consequence, 
\begin{equation}
V(\underline{J}(A,m)) \leq \frac{1}{m^{n^2-1}} \left|A \cap \pi\left(\mathcal{C}_{m,n^2}^*\right)\right| \leq V(\overline{J}(A,m)).
\end{equation}

However, $\lim_{m \rightarrow +\infty}V(\underline{J}(A,m)) = \lim_{m \rightarrow +\infty} V(\overline{J}(A,m))$  because $A$ is Riemann-measurable. 
From the squeeze theorem, it follows that 
\begin{equation}
V(A) = \lim_{m \rightarrow +\infty} \frac{1}{m^{n^2-1}} \left|A \cap \pi\left(\mathcal{C}_{m,n^2}^*\right)\right|.
\end{equation}
Hence, the thesis holds.
\end{proof}

%\begin{theorem}
%If $\mathcal{S}_{\sigma,n}(c)$ is Riemann-measurable in dimension $n^2-1$, then
%\begin{equation}
%\lim_{m\rightarrow +\infty}\frac{|\mathcal{R}_{\sigma,n,m}^*(c)|}{|\mathcal{C}_{m,n^2}^*|}=\rho_{\sigma, n}(c).
%\end{equation}
%\end{theorem}
We are now ready to prove Theorem~\ref{theo:riemann}.
\begin{proof}
Since $\mathcal{S}_{\sigma, n}(c)$ and $\mathcal{C}_{m,n^2}^*$ are subsets of 
$\Delta^{(n^2-1)}$, and since  $\pi$ is bijective, 
\begin{align*}
\pi\left(\mathcal{S}_{\sigma, n}(c)\right) \cap \pi\left(\mathcal{C}_{m,n^2}^*\right) = \pi\left(\mathcal{S}_{\sigma, n}(c) \cap \mathcal{C}_{m,n^2}^*\right)
\end{align*}
Hence, 
\begin{align*}
V(\pi\left(\mathcal{S}_{\sigma, n}(c)\right)) &=\lim_{m \rightarrow +\infty} \frac{1}{m^{n^2-1}} \left| \pi\left(\mathcal{S}_{\sigma, n}(c)\right) \cap \pi\left(\mathcal{C}_{m,n^2}^*\right)\right|\\
&= \lim_{m \rightarrow +\infty} \frac{1}{m^{n^2-1}} \left| \pi\left(\mathcal{S}_{\sigma, n}(c) \cap \mathcal{C}_{m,n^2}^*\right)\right|\\
&= \lim_{m \rightarrow +\infty} \frac{1}{m^{n^2-1}} \left| \pi\left(\mathcal{R}_{\sigma,n,m}^*(c)\right)\right|\\
&= \lim_{m \rightarrow +\infty} \frac{1}{m^{n^2-1}} \left| \mathcal{R}_{\sigma,n,m}^*(c)\right|
\end{align*}
because of Lemma~\ref{lemma:riemann} and Eq.~\ref{eq:pR_n_m}.

However,
\begin{align*}
\pi\left(\Delta^{(n^2-1)}\right) \cap \pi\left(\mathcal{C}_{m,n^2}^*\right) = \pi\left(\mathcal{C}_{m,n^2}^*\right)
\end{align*}
because $\mathcal{C}_{m,n^2}^* \subseteq \Delta^{(n^2-1)}$ and because $\pi$ is bijective.
Thus,
\begin{align*}
V(\pi\left(\Delta^{(n^2-1)}\right)) = \lim_{m \rightarrow +\infty} \frac{1}{m^{n^2-1}} \left| \pi\left(\mathcal{C}_{m,n^2}^*\right)\right|
= \lim_{m \rightarrow +\infty} \frac{1}{m^{n^2-1}} \left| \mathcal{C}_{m,n^2}^*\right|
\end{align*}
because of Lemma~\ref{lemma:riemann}. It follows that
\begin{align*}
\rho_{\sigma, n}(c)=\frac{V(\mathcal{S}_{\sigma, n}(c))}{V(\Delta^{(n^2-1)})}=\frac{\lim_{m \rightarrow +\infty} \frac{1}{m^{n^2-1}} \left| \mathcal{R}_{\sigma,n,m}^*(c)\right|}{\lim_{m \rightarrow +\infty} \frac{1}{m^{n^2-1}} \left| \mathcal{C}_{m,n^2}^*\right|}=\lim_{m \rightarrow +\infty}\frac{\left| \mathcal{R}_{\sigma,n,m}^*(c)\right|}{\left| \mathcal{C}_{m,n^2}^*\right|}.
\end{align*}
Since $\left| \mathcal{C}_{m,n^2}^*\right|=\left| \mathcal{C}_{m,n^2}\right|$, 
\begin{align*}
\rho_{\sigma, n}(c)=\lim_{m \rightarrow +\infty}\frac{\left| \mathcal{R}_{\sigma,n,m}^*(c)\right|}{\left| \mathcal{C}_{m,n^2}^*\right|}
=\lim_{m \rightarrow +\infty}\frac{\left| \mathcal{R}_{\sigma,n,m}(c)\right|}{\left| \mathcal{C}_{m,n^2}\right|}
=\lim_{m \rightarrow +\infty} \varrho_{\sigma, n, m}(c).
\end{align*}
because of Lemma~\ref{lemma:sigma} and Eq.~\ref{eq:varrho}.
\end{proof}

\end{document}